\documentclass[lettersize,journal]{IEEEtran}
\usepackage{amsmath,amsfonts}

\usepackage[linesnumbered,ruled,vlined]{algorithm2e}
\SetKwInput{KwInput}{Input}                
\SetKwInput{KwOutput}{Out}                 
\SetKwFor{KwFor}{for each}{do}{}           
\SetKwInput{KwData}{Global params.}        

\usepackage{graphicx}
\usepackage{tikz}
\usetikzlibrary{arrows.meta, backgrounds}
\usepackage{mwe} 
\usepackage{booktabs} 
\usepackage{multirow} 
\usetikzlibrary{shadows}
\usetikzlibrary{shadows.blur}
\usetikzlibrary{calc,math}
\usepackage{array}
\usepackage{subcaption}
\usepackage{textcomp}
\usepackage{contour}
\usepackage{stfloats}
\usepackage{url}
\usepackage{verbatim} 
\usepackage{bm}
\usepackage{tikz}
\usepackage{soul}
\tikzset{refpath/.style={thick}}
\usetikzlibrary{arrows.meta}
\usetikzlibrary{shapes}
\usetikzlibrary{calc}
\usetikzlibrary{angles,quotes}
\usetikzlibrary{decorations.pathreplacing}
\usetikzlibrary{shapes.geometric}
\usepackage{graphicx}
\usepackage{cite}
\usepackage{color}
\usepackage{xcolor}
\usepackage{acronym}
\usepackage{siunitx}
\providecommand{\qty}{\SI}
\providecommand{\qtyrange}{\SIrange}
\usepackage{adjustbox}

\usepackage{hyperref}
\hypersetup{
    colorlinks=true,
    linkcolor=black,
    filecolor=black,      
    urlcolor=black,
    citecolor=black,
    }

\acrodef{NMPC}{Nonlinear Model Predictive Control}
\acrodef{MPCC}{Model Predictive Contouring Control}
\acrodef{FF-MPCC}{Formation Flight Model Predictive Contouring Control}
\acrodef{UAV}{Uncrewed Aerial Vehicle}
\acrodef{MAPDE}{Mean Absolute Pairwise Distance Error}
\acrodef{SFTE}{Spatial Formation Tracking Error}

\newcommand{\reffig}[1]{Fig.~\ref{#1}}

\newcommand{\refalg}[1]{Alg.~\ref{#1}}
\newcommand{\refsec}[1]{Sec.~\ref{#1}}
\newcommand{\reftab}[1]{Table~\ref{#1}}
\newcommand{\refeq}[1]{\eqref{#1}}
\newcommand{\shapeformat}{\bm{M}}

\newcommand{\relativecoordvec}{\bm{r}_i(\tau)}
\newcommand{\formationheading}{\psi}

\newcommand{\pathparam}{\tau}

\acrodef{uav}[UAV]{Uncrewed Aerial Vehicle}

\usepackage{titlesec}

\titlespacing*{\section}{0pt}{5pt}{1.5pt}
\titlespacing*{\subsection}{0pt}{3pt}{1.5pt}
\titlespacing*{\subsubsection}{0pt}{2pt}{0.5pt}

\usepackage[placement=top,vshift=1,firstpage=true]{background}
\SetBgScale{1.0}
\SetBgContents{\parbox{0.95\textwidth}{\small \begin{center} This work has been submitted to the IEEE Robotics and Automation Letters for possible publication. Copyright may be transferred without notice, after which this version may no longer be accessible. \end{center}}}
\SetBgColor{black}
\SetBgAngle{0}
\SetBgOpacity{1.0}

\begin{document}
\newcommand{\bs}[1]{\bm{#1}}

\title{FF-MPCC: High-speed Agile Formation Flight with Model Predictive Contouring Control}
\vspace{-12pt}

\author{
Aditya Dandwate,
V{\'{i}}t Kr{\'{a}}tk{\'{y}},
Parakh M. Gupta,
Martin Saska,
Robert P{\v{e}}ni{\v{c}}ka
\thanks{The authors are with the Multi-robot Systems Group, Faculty of Electrical
Engineering, Czech Technical University in Prague, Czech Republic (\protect\url{http://mrs.fel.cvut.cz/}). 
This work has been supported by the Czech Science Foundation (GA\v{C}R) under research project No. 23-06162M, by the European Union under the project Robotics and advanced industrial production (reg. no. CZ.02.01.01/00  /22\_008/0004590), and by CTU grant no. SGS26/077/OHK3/1T/13.}
}

\maketitle

\begin{abstract}
Flying in a prescribed formation in an agile manner remains a challenging problem in the field of \acp{UAV}, particularly when following highly-demanding trajectories that require flight at platform limits. 
We address this problem by proposing a novel decentralized approach to formation flight along a given path that integrates formation maintenance into the \ac{MPCC} framework allowing \acp{UAV} to adapt their progression along complex paths while respecting individual dynamic constraints and maintaining desired formation. 
To this end, we introduce a novel reparametrization and synchronization method for dynamic formation geometries together with a decentralized approach to determine the desired positions for the individual \acp{UAV}. The proposed approach allows the formation to coordinate high-speed path following without compromising the formation integrity. 
The proposed approach is validated through extensive simulation and real-world experiments involving scenarios with varying complexity of paths and changes of required formation shape on the fly. 
In comparison to time-parameterized trajectory tracking, we demonstrate improved formation maintenance by 65\% in high-speed flight with velocities up to \SI{21}{\meter\per\second}, while achieving comparable times required to reach the goal.

\end{abstract}

\vspace{-0.7em}
\section*{Supplementary Material}
{\footnotesize
\vspace{-0.3em}
\noindent \textbf{Video:} \url{https://youtu.be/MCOkhx0mBJc} \\
\vspace{-1.0em}
}

\acresetall
\vspace{-0.2cm}
\section{Introduction}
Teams of multi-rotor \acp{UAV} operating in formation enable applications that are difficult or impossible for a single vehicle, including coordinated inspections~\cite{petracek2024historicpreservation}, environmental monitoring~\cite{brust2015envmapping} and cooperative transportation~\cite{ccari2023novelformationtransport}, where formations provide increased sensing coverage, robustness, and task execution efficiency. However, achieving these benefits in real-world deployments often requires reliable formation flight in high-speed scenarios under aggressive maneuvers, changing mission objectives, and operation in dynamic environments.

Formation flight under these conditions is considerably more challenging than single-\ac{UAV} navigation.
A planned trajectory that is dynamically feasible for the formation center is not necessarily feasible for all \acp{UAV} since individual formation members experience different path curvatures, and therefore require different velocities and accelerations.
Consequently, existing methods typically require dynamically feasible trajectories to be generated simultaneously for every UAV considering the formation constraints~\cite{saska2020formation, quan2023formationFlightInDenseEnvironments}. 
However, due to their computationally-demanding nature, such approaches limit the reactivity and only allow for low-rate adaptation to disturbance or communication failures.
An alternative approach that plans trajectories with significantly tighter dynamic constraints than the actual vehicle limits to ensure feasibility for all formation members yields overly conservative trajectories that substantially reduce flight speed and overall flight efficiency.

\ac{MPCC}~\cite{lam2010mpcc} addresses similar challenges in single-UAV flight by optimizing progression along a geometric path online instead of tracking a precomputed time-parametrized trajectory. 
Although this enables adaptation to disturbances and dynamic environments, as demonstrated in aggressive single-UAV flight~\cite{romero2022mpcc}, extending \ac{MPCC} to decentralized formation flight remains challenging. 


\usetikzlibrary{positioning}
\contourlength{0.9pt}

\tikzset{
    time_label/.style={
        font=\bfseries\footnotesize,
        text=white,
        align=center
    }
}

\newcommand{\outlined}[1]{%
    \contour{black}{#1}%
}

\begin{figure}[!t]
    \centering

    \begin{tikzpicture}
        \definecolor{org}{RGB}{0,0,0}

        \node[anchor=south west, inner sep=0] (image) at (0,0) {%
            \adjincludegraphics[
                width=\columnwidth,
                trim={0pt 0pt 0pt 100pt},
                clip
            ]{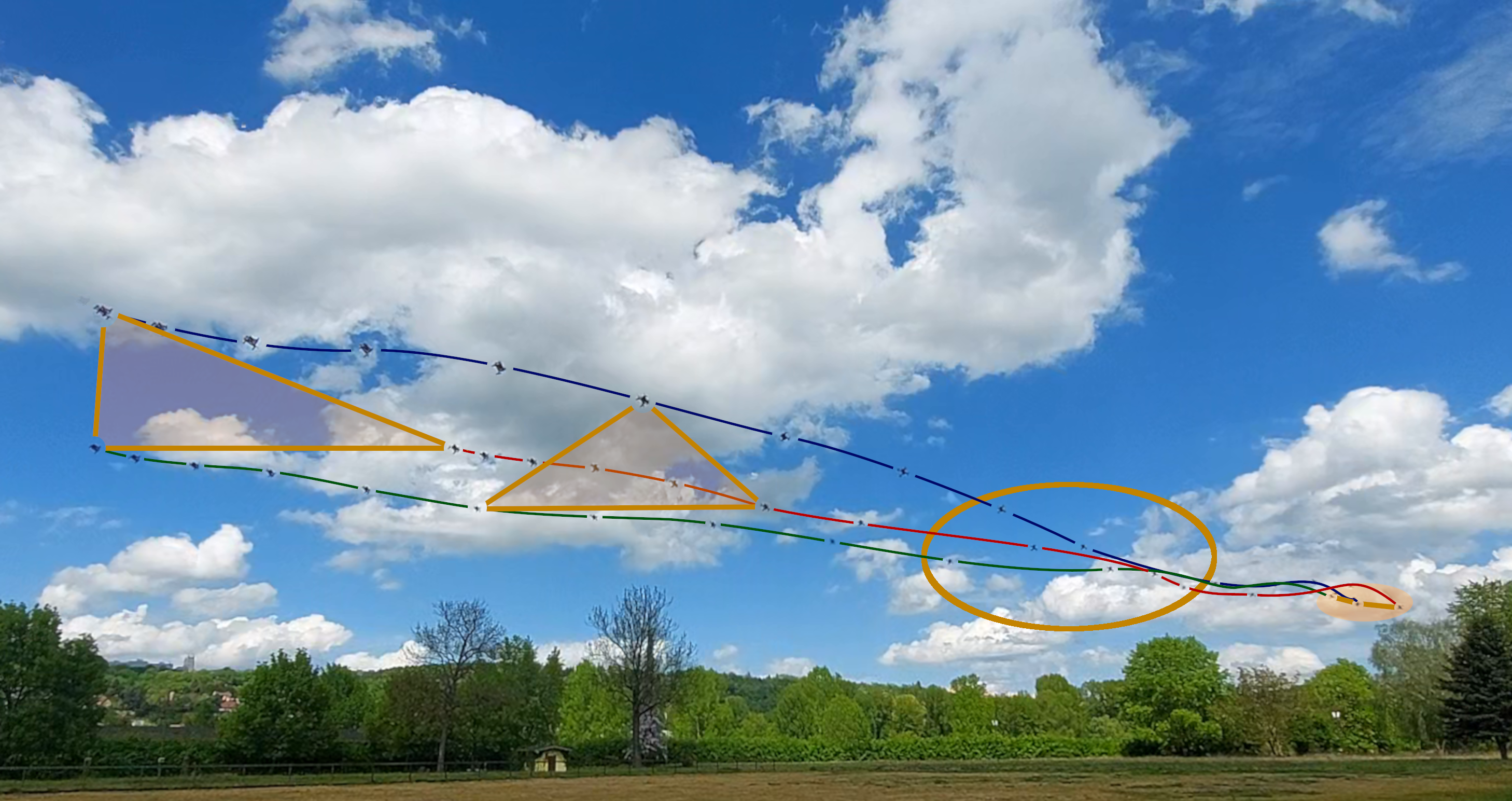}%
        };

        \begin{scope}[x={(image.south east)}, y={(image.north west)}]

            \node[time_label] at (0.1,0.77) {\outlined{0 s}};
            \node[time_label] at (0.43,0.66) {\outlined{2.3 s}};
            \node[time_label] at (0.71,0.53) {\outlined{\shortstack{Transition}}};
            \node[time_label] at (0.90,0.39) {\outlined{6.1 s}};

        \draw[
            white,
            line width=2pt,
            line cap=round
        ] (0.14,0.07) -- (0.91,0.07);
        
        \foreach \x in {0.14,0.91}
            \draw[white,line width=2pt,line cap=round]
                (\x,0.055)--(\x,0.085);
        
        \node[time_label] at (0.525,0.11)
        {\outlined{\SI{60}{\meter}}};
        
        \end{scope}
    \end{tikzpicture}
    \vspace{-0.5cm}
    \caption{
        Deployment of the proposed approach in a real-world scenario with
        three \acp{UAV} navigating along a straight line with transition from triangular to straight line formation with speeds up
        to \SI{21}{\meter\per\second}.
    }
    \label{fig:intro}
    \vspace{-0.7cm}
\end{figure}
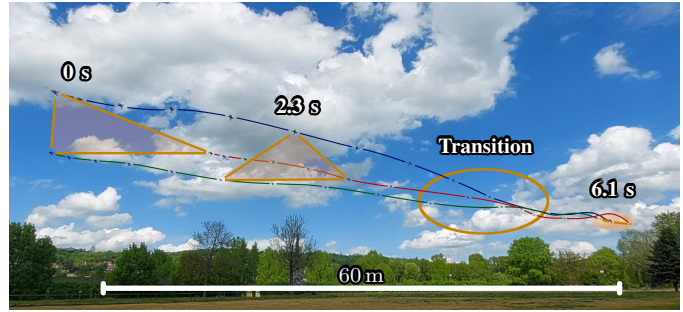
Existing formation control methods primarily focus on maintaining cohesion and collision avoidance, while often sacrificing the time-efficiency of the operation\cite{saska2020formation, quan2023formationFlightInDenseEnvironments, zhou2025reformation}. 
The methods where time optimality is a primary objective typically face the same issues with pre-plannning as their single-UAV counterparts along with the additional overhead of maintaining feasible trajectories for all UAVs in the formation~\cite{kratky2025catora}.
On the other hand, the existing \ac{MPCC}-based formation approaches ignore or relax strict formation constraints in favour of time optimality and general swarming \cite{deng2026mpccFormation, guan2025learningDmpccWithOrca, guevara2024modelpredictivecontouringcontrol}. 
Consequently, none of the existing approaches is able to provide decentralized agile formation flight that combines online trajectory optimization with preservation of prescribed, dynamically evolving formation shapes.

To address this problem, we propose \emph{\ac{FF-MPCC}} decentralized control framework for aggressive \ac{UAV} formation flight.
\ac{FF-MPCC} jointly optimizes progression along the path and adherence to the desired formation geometry within a single optimization problem. 
The synchronization among \acp{UAV} is achieved through a heading-aware formation representation that parameterizes continuously evolving formation geometries by path progression, combined with a distributed rigid-body fitting procedure that constructs a formation-consistent reference from neighboring \acp{UAV} states. 
This enables each \ac{UAV} to continuously minimize formation errors and adapt to disturbances while independently optimizing its own motion and progress along the path. 

The proposed approach is evaluated through extensive simulations against a conventional planning-and-control pipeline based on preplanned minimum-time trajectories and \ac{NMPC}-based trajectory tracking, and is further validated in real-world experiments (see \reffig{fig:intro}) at velocities of up to \SI{21}{\meter\per\second} and acceleration limits of \SI{30}{\meter\per\second\squared}. 
Compared with the baseline, \ac{FF-MPCC} improves formation accuracy by $65\%$ while maintaining comparable or better flight times.
The results demonstrate that fast agile formation flight can be achieved in a decentralized manner by imposing the formation constraints directly at the control level, using only low-bandwidth communication of UAV states and no preplanned dynamically feasible trajectories.

\section{Related Works}

The problem addressed in this work lies at the intersection of single-UAV agile flight and multi-UAV formation control. 
Both fields have experienced significant advances towards increase in speed of the motion, reliability, and scalability in recent years. 
However, they have evolved largely separately.
Methods for agile single-UAV flight typically do not consider formation constraints, while formation-flight methods often prioritize cohesion, collision avoidance, or reconfiguration over time-optimal progression towards goal region.

In the field of single-UAV agile flight, Model Predictive Control (MPC) based algorithms are widely used due to its ability to explicitly handle system constraints while optimizing future control actions~\cite{nguyen2021mpcsurvey, deihl2007shootingnodes}.
In particular, NMPC has become increasingly popular for agile flight, as it allows the nonlinear vehicle dynamics and actuator limits to be considered directly in the optimization problem~\cite{gupta2025lolNmpc, sun2022comparative, gomaa2022computationally}.
This makes NMPC well-suited for accurate trajectory tracking under aggressive manoeuvres while staying within actuator constraints.
However, NMPC-based approaches usually rely on dynamically-feasible reference trajectories generated by a separate planner, commonly using time-optimal planning algorithms~\cite{penicka2022mintimeplanning,teissing2024PMM}.
Although these planners can produce highly dynamic motions, the separation between planning and control becomes limiting when the reference trajectory is difficult to track, or needs to be replanned on the flight. 

MPCC addresses this limitation by optimizing progression along a geometric path while minimizing contouring and lag errors, instead of tracking fully time-parametrized reference trajectory. 
Thus, it removes the need to prescribe the temporal evolution of the reference in advance and allows the controller to determine the progression along the path online. 
MPCC has already been applied to quadrotors and other robotic systems~\cite{ji2021cmpcccorridorbasedmodelpredictive,guevara2024modelpredictivecontouringcontrol, Liniger_2014}, and has demonstrated strong performance in agile flying and drone racing \cite{romero2022mpcc,romero2022replanning}. 
However, these works focus on single-vehicle operation only.

Formation flight has been extensively studied in the context of multi-UAV systems.
Existing methods typically focus on formation cohesion and reshaping~\cite{kratky2025catora}, obstacle avoidance~\cite{wang2021uav, seo2017collision}, communication constraints, and decentralized coordination.
A large body of formation-flight research considers navigation in cluttered or obstacle-rich environments~\cite{quan2023formationFlightInDenseEnvironments, zhou2025reformation, zhang2025agile},
These approaches demonstrate the importance of adapting the formation to the environment, but generally do not focus on time-optimal progression along a prescribed trajectory.

Several works have specifically explored MPCC or MPCC-inspired approaches for multi-robot systems and formation flight. 
The authors of \cite{guan2025learningDmpccWithOrca} propose a neural network to learn time optimal allocations to facilitate online re-planning, but considers only mutual collision and do not enforce any other mean of coordination.
Similarly, time-optimal swarm trajectory generation has been investigated in~\cite{pan2025tstar}, where MPCC is used to improve the temporal efficiency of swarm motion. 
Bio-inspired MPCC formulations for fixed-wing swarms have also been explored~\cite{deng2026mpccFormation}, although their assumptions and vehicle dynamics differ from the agile multirotor UAVs. 
While these methods improve swarm performance, and address general time optimal trajectory planning, they do not consider formation geometry constraints.

This work bridges the gap between agile MPCC-based flight and formation-constrained multi-UAV flight by integrating formation constraints directly into the nonlinear MPCC formulation, combined with an online distributed formation fitting that provides every UAV with its formation-consistent target position from the current and predicted states of its neighbours. 
This enables coordinated UAV teams to navigate aggressively along a shared path while respecting both vehicle dynamics and formation geometry.

\section{Methodology}

This section first introduces the quadrotor model, followed by a description of the proposed \ac{FF-MPCC} optimal control problem formulation, reference formation path generation, and formation coordination approach. 

\subsection{Quadrotor Dynamics}
\label{subsec:quadrotor_dynamics}

The quadrotor state can be described by its position $\bm{p} \in \mathbb{R}^3$, \ orientation $\bm{q} \in \mathbb{SO}(3)$, linear velocity $\bm{v} \in \mathbb{R}^3$ and angular velocity in body frame $\bm{\omega} \in \mathbb{R}^3$. 
The input to the dynamics~\eqref{eq:dynamic_model} is the collective thrust of the quadrotor in body frame $\bm{f}_T = \begin{bmatrix} 0 & 0 & f \end{bmatrix}^T$ and body torques $\bm{\tau}$.
The dynamics in world-frame can be described by differential equations
\begin{equation}
\label{eq:dynamic_model}
\begin{array}{cc}
    \begin{split}
        \dot{\bm{p}} &= \bm{v}\text{,}
    \end{split} 
    & 
    \begin{split}
        \dot{\bm{v}} &= \frac{1}{m}\bm{R}(\bm{q})(\bm{f}_T + \bm{f}_D) + \bm{g} \text{,}
    \end{split} 
    \\[2ex]
    \begin{split}
        \dot{\bm{q}} &= \frac{1}{2} \bm{q} \odot \begin{bmatrix} 0 \\ \bm{\omega} \end{bmatrix}\text{,}
    \end{split}  
    & 
    \begin{split}
        \dot{\bm{\omega}} &= \bm{J}^{-1}(\bm{\tau} - \bm{\omega} \times \bm{J}\bm{\omega})\text{,}
    \end{split}
\end{array}
\end{equation}
where $\odot$ represents the quaternion multiplication, $\bm{R}(\bm{q})$ the rotation matrix of $\bm{q}$, $m$ the quadrotor mass, $\bm{J}$ its inertia, $\bm{g}$ the gravity vector, and $\bm{f}_D$ the drag force.

We model the drag $\bm{f}_D$ as a linear function of the velocity in the body frame, such that $\bm{f}_D = -\bm{k}_{v}\bm{v}_{\mathcal{B}}$, where $\bm{v}_{\mathcal{B}} = \bm{R}^{T}(\bm{q})\bm{v}$ is the velocity in the body frame and $\bm{k}_{v} = \begin{bmatrix} k_{vx} & k_{vy} & k_{vz} \end{bmatrix}$ is the vector of drag coefficients.

The collective thrust $f$ in $\bm{f}_T$ and the body torques $\bm{\tau}$ can be calculated from the individual rotor thrusts $\bm{f} = [f_1, f_2, f_3, f_4]$  of motor with propeller $i \in \{1, 2, 3, 4\}$ as:
\begin{equation}
\label{eq: thrust_model}
\begin{array}{ccc}
    f = \sum_{i=1}^4 f_i\text{, } 
    & \quad &
    \bm{\tau} = \begin{bmatrix}
        \frac{l}{\sqrt{2}}(f_1 - f_2 - f_3 + f_4) \\ 
        \frac{l}{\sqrt{2}}(- f_1 - f_2 + f_3 + f_4) \\ 
        \kappa(f_1 - f_2 + f_3 - f_4) 
    \end{bmatrix}
    \text{,}
\end{array}
\end{equation}
where $l$ is the distance from the center of mass of a symmetric quadrotor to each motor and $\kappa$ is the torque coefficient of the motor with propeller.

\subsection{Formation Flight Model Predictive Contouring Control Formulation}
\label{subsec:formation_formulation}

In contrast to the classical \ac{NMPC}~\cite{gupta2025lolNmpc}, which tracks a full-state reference trajectory explicitly parameterized in time, the \ac{FF-MPCC} framework, similarly to \ac{MPCC}, tracks a geometric path parameterized by its arc length instead. 
The time parametrization along the path is not prescribed a priori but is determined online by the \ac{FF-MPCC} solver.
As per standard \ac{MPCC}\cite{romero2022mpcc} frameworks, the cumulative arc length of the reference path (or progress) is denoted by $\theta$, and the arc length at time-step $k$ is denoted by $\theta_k$. 
The \ac{FF-MPCC} cost function comprises  four principal terms: a \emph{lag error} term $\bm{e}_l(\theta_k)$ penalizing deviation along the direction of travel, a \emph{contour error} term $\bm{e}_c(\theta_k)$ penalizing lateral deviation from the path, a \emph{progress maximization} term $v_\theta$ that encourages fast forward motion, and a \emph{formation error} term $\bm{e}_f(\theta_k)$ penalizing deviation from the desired geometric relationship among drones in the formation. 
By optimizing this cost function at each time step, \ac{FF-MPCC} dynamically balances the path following accuracy against forward advancement along the path and the precision of keeping the given formation shape.

The reference 3D path parametrized by its arc length is $\bm{p}_r(\theta) = [x_r(\theta), y_r(\theta), z_r(\theta)]^T$ and the actual spatial position of the drone at time-step $k$ is denoted by $\bm{p}_k = [x_k, y_k, z_k]^T$.
Let $\bm{t}(\theta_k) \in \mathbb{R}^3$ be the tangent of $\bm{p}_r(\theta_k)$ evaluated at $\theta_k$. 
The position error at time-step $k$ can be then computed as $\bm{e}(\theta_k) = \bm{p}_k - \bm{p}_r(\theta_k)$, and can be decomposed into the contour error $\bm{e}_c(\theta_k)$ and lag error $\bm{e}_l(\theta_k)$ such that 
\begin{equation}\label{eq:lag_contour_decomposition}
\bm{e}(\theta_k) = \bm{e}_l(\theta_k) + \bm{e}_c(\theta_k)\text{.}
\end{equation}
The lag error is approximated by projecting $\bm{e}(\theta_k)$ onto $\bm{t}(\theta_k)$,
\begin{equation}\label{eq:lag_tangent_projection}
\bm{e}_l(\theta_k) = (\bm{t}(\theta_k)^T \bm{e}(\theta_k)) \bm{t}(\theta_k) \text{.}
\end{equation}
The arc-length parametrization of $\bm{p}_r(\theta)$, which by definition has a unit norm of the tangent $\bm{t}(\theta_k)$ for any $\theta_k$, allows the lag error to be expressed as a scalar $e_l(\theta_k)$.
The contour error is computed as a difference between the position error $\bm{e}(\theta_k)$ and the lag error $\bm{e}_l(\theta_k)$, i.e., $\bm{e}_c(\theta_k) = \bm{e}(\theta_k) - \bm{e}_l(\theta_k)$.
This means the contour error is the projection of the position error onto the plane orthogonal to the tangent vector $\bm{t}(\theta_k)$ in $\bm{p}_k$.

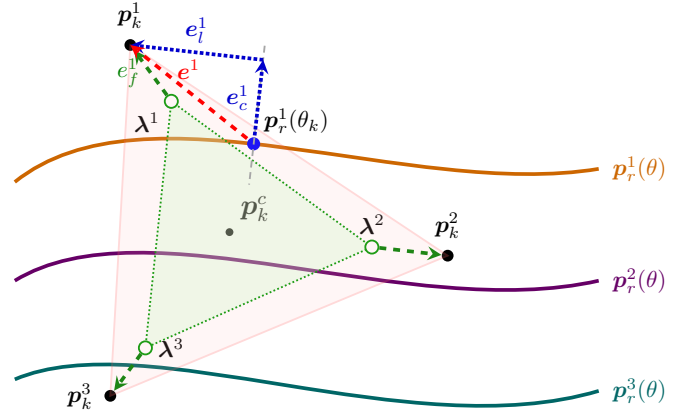
\begin{figure}[h!]
    \centering
    \resizebox{\columnwidth}{!}{\begin{tikzpicture}[
        scale=2.5,
        >=stealth,
        path1/.style={ultra thick, orange!80!black},
        path2/.style={ultra thick, violet!80!black},
        path3/.style={ultra thick, teal!80!black},
        drone/.style={circle,fill=black,inner sep=1.8pt},
        error/.style={->,ultra thick,red, dashed},
        comp/.style={->, ultra thick, blue!80!black, densely dotted},
        formerror/.style={->,ultra thick, green!50!black, dashed},
        progress/.style={circle,fill=blue,inner sep=2.0pt},
        target/.style={
            circle,
            draw=green!60!black,
            fill=white,
            thick,
            inner sep=2pt
        }
    ]
    
    \draw[path1] (0,1.6) .. controls (0.8,2.24) and (2.4,1.49) .. (3.55,1.68)
        node[pos=0.45,progress] (pd1) {}
        node[pos=0.46] (pd1_next) {} 
        node[right=2pt, orange!80!black] {$\boldsymbol{p}^1_r(\theta)$};

    \node[above right=1.5pt] at (pd1) {$\bs{p}_r^1(\theta_k)$};
    
    \draw[path2] (0,1.0) .. controls (0.8,1.49) and (2.4,0.74) .. (3.55,1.0)
        node[pos=0.65,coordinate] (pd2) {}
        node[pos=0.66] (pd2_next) {} 
        node[right=2pt, violet!80!black] {$\boldsymbol{p}^2_r(\theta)$};
    
    \draw[path3] (0,0.4) .. controls (0.8,0.74) and (2.4,-0.01) .. (3.55,0.33)
        node[pos=0.35,coordinate] (pd3) {}
        node[pos=0.36] (pd3_next) {}
        node[right=2pt, teal!80!black] {$\boldsymbol{p}^3_r(\theta)$};
    
    \coordinate (pc1) at ($(pd1)+(-0.75,0.6)$);
    \coordinate (pc2) at ($(pd2)+(0.4,0.15)$);
    \coordinate (pc3) at ($(pd3)+(-0.5,-0.15)$);
    
    \node[drone] (c1) at (pc1) {};
    \node[drone] (c2) at (pc2) {};
    \node[drone] (c3) at (pc3) {};
    
    \node[above=4pt] at (pc1) {$\boldsymbol{p}_{k}^{1}$};
    \node[above=4pt] at (pc2) {$\boldsymbol{p}_{k}^{2}$};
    \node[left=4pt] at (pc3) {$\boldsymbol{p}_{k}^{3}$};
    
    \coordinate (centroid) at ($1/3*(pc1) + 1/3*(pc2) + 1/3*(pc3)$);

    \node[circle, fill=black, inner sep=1.2pt, label={above right:\large $\boldsymbol{p}_k^c$}] at (centroid) {};

    \pgfmathsetmacro{\d}{0.87} 
    \pgfmathsetmacro{\phiAngle}{-6} 
    
    \coordinate (pf1) at ($(centroid)+({\d*cos(\phiAngle+120)},{\d*sin(\phiAngle+120)})$);
    \coordinate (pf2) at ($(centroid)+({\d*cos(\phiAngle)},{\d*sin(\phiAngle)})$);
    \coordinate (pf3) at ($(centroid)+({\d*cos(\phiAngle+240)},{\d*sin(\phiAngle+240)})$);
    
    \draw[green!60!black,densely dotted, thick, fill=green!20, fill opacity=0.3] (pf1)--(pf2)--(pf3)--cycle;
    
    \node[target] (v1) at (pf1) {}; 
    \node[target] (v2) at (pf2) {}; 
    \node[target] (v3) at (pf3) {};

    \node[below left=2pt]  at (v1) {$\bs{\lambda}^{1}$};
    \node[above=2.3pt] at (v2) {$\bs{\lambda}^{2}$};
    \node[right=2pt] at (v3) {$\bs{\lambda}^{3}$}; 

    \draw[formerror] (v1) -- (c1) node[midway, below=2pt, left=0.5pt] {$e_f^1$};
    \draw[formerror] (v2) -- (c2) node[midway, below=2pt] {};
    \draw[formerror] (v3) -- (c3) node[midway, above=2pt] {};

    
    \draw[
    red!60,
    thick,
    fill=red!30,
    fill opacity=0.12,
    draw opacity=0.25
    ] (pc1) -- (pc2) -- (pc3) -- cycle;
    
    
    \path let
        \p1=(pd1),
        \p2=(pd1_next),
        \n1={atan2(\y2-\y1,\x2-\x1)}
    in
    {
        \pgfextra{\xdef\tangentangle{\n1}}
    };
    
    \coordinate (normalTop1) at ($(pd1)+(\tangentangle+90:0.6)$);
    
    \draw[densely dashed, thick, gray!70]
        ($(pd1)+(\tangentangle+90:0.6)$)
        --
        ($(pd1)+(\tangentangle-90:0.25)$);
    
    
    \coordinate (p_contour) at ($(normalTop1)!(pc1)!(pd1)$);
    
    \draw[comp]
        (pd1) -- (p_contour)
        node[midway,left] {$\boldsymbol{e}_{c}^1$};
    
    \draw[comp]
        (p_contour) -- (pc1)
        node[midway,above] {$\boldsymbol{e}_{l}^1$};

    \draw[error] (pd1) -- (pc1) node[midway, above=1.5pt] {$\boldsymbol{e}^1$};
    
\end{tikzpicture}}
    \vspace{-2.5em}
    \caption{Geometric error formulation for a three-UAV formation at time-step $k$. The current position of UAV $D_i$ is denoted by $\bs{p}_k^i$, with $\bs{p}_k^c$ representing the formation center. The target geometry positions are defined by $\bs{\lambda}^i$, from which the formation errors $\bs{e}_f^i$ are calculated. The total tracking error $\bs{e}^1$ is the vector sum of the contour error $\bs{e}_c^1$ and lag error $\bs{e}_l^1$, which are defined relative to the reference position $\bs{p}_r^1(\theta_k)$ on the arc-length parametrized reference path. 
    \label{fig:formulation_tikz}}
    \vspace{-1em}
\end{figure}

Arc-length parametrized paths $\{p^1_r(\theta),\dots,p^N_r(\theta)\}$ for all \acp{UAV} $D_i,\text{ }i \in \{1, \dots, N\}$ are generated as described in \refsec{subsec:formation_path_generation}.
The paths are generated such that for every point on the formation-center path, the corresponding points on the individual \ac{UAV} paths form exactly the desired formation shape.
However, the \ac{FF-MPCC} solvers of individual \acp{UAV} are decoupled, and thus have the freedom to track their paths of different lengths with distinct velocity profiles, which might result in non-cohesive formations.
\ac{FF-MPCC} resolves this by interlinking the \acp{UAV} in a decentralized fashion: at each control iteration, ego-UAV $D_i$ receives the current position and the predicted positions from the \ac{FF-MPCC} solution of all other-\acp{UAV} $\mathcal{D}_{j}, j \in \{1,\dots,N\} \setminus \{i\}$.
To enforce formation geometry, a formation target position $\bm{\lambda}^i_k \in \mathbb{R}^3$ is defined for each $D_i$ and time step $k$ as a world-frame position that best preserves the prescribed formation shape given the actual or predicted positions of the ego-\ac{UAV} $D_i$ and all other-\acp{UAV} $\mathcal{D}_{j}$ at time step $k$.
To this end, the target position is obtained by rigid body fitting of the reference formation shape to the current and predicted \acp{UAV}' positions.
The fitting procedure is described in more detail in \refsec{subsec:formation_coordination}.
The formation target position $\bm{\lambda}_k$ is then used to define a \emph{formation error} $\bm{e}_{f,k}$ as the Euclidean distance between its actual position $\bm{p}_k$ and its corresponding shape fitting target position $\bm{\lambda}_k$ with
\begin{equation}
\label{eq:formation_error_basic}
\bm{e}_{f,k} = \left\lVert \bm{p}_{k} - \bm{\lambda}_{k} \right\rVert_2.
\end{equation}
The formation error $\bm{e}_{f,k}$ is then minimized as part of the \ac{FF-MPCC} optimization problem, which encourages the UAV to synchronize in order to maintain the desired formation geometry while tracking its reference path and maximizing progress along it.
All the formation, contour and lag errors are illustrated in a 2D example in \reffig{fig:formulation_tikz} for a triangular formation of three \acp{UAV}.

In parallel with path tracking accuracy and the maintenance of the formation shape, \ac{FF-MPCC} maximizes the progress along the reference path. 
Following the single-drone \ac{MPCC} formulation proposed in \cite{romero2022mpcc}, the progress-maximization objective is encoded directly in the cost function by maximizing the arc-length speed $v_{\theta,k}$. 
This term incentivizes the drone to advance along the path, while the lag, contour, and formation terms jointly constrain the deviations from the reference path and formation shape.  

The state of the modelled \ac{UAV} dynamics in \eqref{eq:dynamic_model} is augmented by the single rotor thrusts $\bm{f}_k$ (where $k$ denotes the time-step), the path progress $\theta_k$, and the virtual speed $v_{\theta,k}$.
To enforce limits on the rate of change of the thrust and virtual speed, thus preventing noisy control inputs, the progress acceleration $\Delta v_{\theta}$ and thrust acceleration $\Delta \bm{f}$ are chosen as the control inputs. 
The resulting augmented state space $\bm{x}$ is
\begin{equation}
\label{eq:augmented_state_space}
    \bm{x} = \begin{bmatrix} \bm{p}^T \quad \bm{q}^T \quad \bm{v}^T \quad \bm{w}^T \quad \bm{f}^T \quad \theta \quad v_\theta \end{bmatrix}^T \text{,}
\end{equation}
the control input with progress acceleration and thrust acceleration, respectively, is
\begin{equation}
\label{eq:input_space}
    \bm{u} = \begin{bmatrix} \Delta v_{\theta} \quad \Delta \bm{f}^T  \end{bmatrix}^T \text{,}
\end{equation}
and the dynamics of the augmented states are
\begin{equation}
\begin{aligned}
\bs{f}_{k+1} &= \bs{f}_k + \Delta \bs{f}_k \Delta t, \\
\theta_{k+1} &= \theta_k + v_{\theta, k}  \Delta t, \\
    v_{\theta, k+1} &= v_{\theta, k} + \Delta v_{\theta, k} \Delta t.
\end{aligned}
\end{equation}
The full \ac{FF-MPCC} optimal control problem with a finite prediction horizon~$N$ is then formulated as follows:
\begin{small}
\label{ff_mpcc_ocp}
\begin{subequations}
\begin{align}
\bm{u}^* = \underset{\bm{u}}{\arg\min} \;
& \sum_{k=0}^{N} 
\Big(
    \| \bm{e}_l(\theta_k) \|^2_{Q_l}
  + \| \bm{e}_c(\theta_k) \|^2_{Q_c}
  + \| \bm{e}_{f,k} \|^2_{Q_{f}} \nonumber \\
  & \quad
  - \mu v_{\theta,k} 
  + \| \bm{\omega}_k \|^2_{Q_{\bm{\omega}}}  
  + \| \Delta v_{\theta,k} \|^2_{R_{\Delta v}} \nonumber \\
  & \quad
  + \| \Delta \bm{f}_k \|^2_{R_{\Delta \bm{f}}}
\Big) \label{eq:ff_mpcc_ocp_objective} 
\\[6pt]
\text{s. t. } & \bm{x}_0 = \bm{x}, \label{ff_mpcc_ocp_init} \\[3pt]
& \bm{x}_{k+1} = f(\bm{x}_k, \bm{u}_k),  \label{ff_mpcc_ocp_dynamics}  \\[3pt]
& \bm{\omega}_{\min} \leq \bm{\omega}_k \leq \bm{\omega}_{\max}, \label{ff_mpcc_ocp_omega} \\
& \bm{f}_{\min} \leq \bm{f}_k \leq \bm{f}_{\max}, \label{ff_mpcc_ocp_f} \\[3pt]
& 0 \leq v_{\theta,k} \leq v_{\theta,\max}, \label{ff_mpcc_ocp_vtheta} \\[3pt]
& \Delta v_{\theta,\min} \leq \Delta v_{\theta,k} \leq \Delta v_{\theta,\max}, \label{ff_mpcc_ocp_deltavtheta} \\[3pt]
& \Delta \bm{f}_{\min} \leq \Delta \bm{f}_k \leq \Delta \bm{f}_{\max} \label{ff_mpcc_ocp_deltaf} .
\end{align}
\end{subequations}
\end{small}

The objective~\eqref{eq:ff_mpcc_ocp_objective} minimizes the lag error, contour error, and formation error, penalized by the positive semi-definite weight matrices $\bm{Q}_l$, $\bm{Q}_c$, and $\bm{Q}_{f}$, respectively, while maximizing progress through the negative term $-\mu v_{\theta,k}$. 
Additionally, the angular velocity $\bm{\omega}_k$, the progress speed change $\Delta v_{\theta,k}$, and the thrust changes $\Delta \bm{f}_k$ are penalized with weights $\bm{Q}_{\bm{\omega}}$, $\bm{R}_{\Delta v}$, and $\bm{R}_{\Delta f}$, respectively, to promote smooth control inputs and avoid aggressive bang-bang behavior. 
The constraints~\eqref{ff_mpcc_ocp_init}--\eqref{ff_mpcc_ocp_deltaf} enforce the system initial state~\eqref{ff_mpcc_ocp_init} and its dynamics~\eqref{ff_mpcc_ocp_dynamics}; and constrain angular velocity~\eqref{ff_mpcc_ocp_omega}, thrust~\eqref{ff_mpcc_ocp_f}, progress speed~\eqref{ff_mpcc_ocp_vtheta}, and the rates of change of progress speed~\eqref{ff_mpcc_ocp_deltavtheta} and thrust~\eqref{ff_mpcc_ocp_deltaf}.

\subsection{Reference Formation Trajectory Generation}
\label{subsec:formation_path_generation}

The reference trajectory of the formation is composed of two coupled components: the global trajectory of the formation center and a time-varying formation geometry that specifies the relative positions of the UAVs. 
Both components are synchronized through a path parameter $\pathparam \in [0,\pathparam_{\max}]$, where each value of $\pathparam$ uniquely identifies a point on the center trajectory and the corresponding formation geometry.

\begin{figure}[h!]
    \centering
    \resizebox{\columnwidth}{!}{`   \begin{tikzpicture}[x=13cm, y=1.8cm, >=Stealth]

    \pgfmathsetmacro{\L}{1.7}           
    \pgfmathsetmacro{\LineL}{\L*0.5}    
    
    \pgfmathsetmacro{\TriWidth}{0.045}  
    
    \pgfmathsetmacro{\tA}{0.20}         
    \pgfmathsetmacro{\tB}{0.4}         
    \pgfmathsetmacro{\tC}{0.60}         
    \pgfmathsetmacro{\tD}{0.80}         

    \pgfmathsetmacro{\Ymid}{1.5}
    \pgfmathsetmacro{\Ytop}{\Ymid + \L/2}
    \pgfmathsetmacro{\Ybot}{\Ymid - \L/2}
    
    \pgfmathsetmacro{\YlineTop}{\Ymid + \LineL/2}
    \pgfmathsetmacro{\YlineBot}{\Ymid - \LineL/2}

    \definecolor{staticColor}{RGB}{230, 242, 250}   
    \definecolor{transColor}{RGB}{254, 243, 226}    
    \definecolor{lineColor}{RGB}{45, 95, 140}       
    \definecolor{boundaryColor}{RGB}{180, 180, 180} 
    \definecolor{droneColor}{RGB}{230, 30, 30}      

    \begin{scope}[on background layer]
        \fill[staticColor] (0.0, 0.4) rectangle (\tA, 2.6); 
        \fill[transColor]  (\tA, 0.4) rectangle (\tB, 2.6); 
        \fill[staticColor] (\tB, 0.4) rectangle (\tC, 2.6); 
        \fill[transColor]  (\tC, 0.4) rectangle (\tD, 2.6); 
        \fill[staticColor] (\tD, 0.4) rectangle (1.0, 2.6); 
    \end{scope}

    \draw[boundaryColor, dashed, line width=1pt] (\tA, 0.4) -- (\tA, 2.6);
    \draw[boundaryColor, dashed, line width=1pt] (\tB, 0.4) -- (\tB, 2.6);
    \draw[boundaryColor, dashed, line width=1pt] (\tC, 0.4) -- (\tC, 2.6);
    \draw[boundaryColor, dashed, line width=1pt] (\tD, 0.4) -- (\tD, 2.6);

    \draw[thick, lineColor] (-0.02, \Ymid) -- (1.02, \Ymid) node[right, black, font=\small] {A};

    \draw[very thick, lineColor] 
        (-0.02, \Ytop) -- (\tA, \Ytop) -- (\tB, \YlineTop) -- (\tC, \YlineTop) -- (\tD, \Ytop) -- (1.02, \Ytop)
        node[right, black, font=\small] {B};

    \draw[very thick, lineColor] 
        (-0.02, \Ybot) -- (\tA, \Ybot) -- (\tB, \YlineBot) -- (\tC, \YlineBot) -- (\tD, \Ybot) -- (1.02, \Ybot)
        node[right, black, font=\small] {C};

    \draw[thick, ->] (-0.02, 0.4) -- (1.05, 0.4) node[right, font=\huge] {$\tau_{\text{norm}}$};
    \foreach \x in {0.0, \tA, \tB, \tC, \tD, 1.0} {
        \draw[thick] (\x, 0.43) -- (\x, 0.37);
        \node[below=3pt, font=\small\bfseries] at (\x, 0.4) {\x};
    }

    \node[above=4pt, font=\large\bfseries, color=lineColor] at ({0.5*\tA}, 2.6) {Static};
    \node[above=4pt, font=\large\bfseries, color=orange!80!black] at ({0.5*(\tA+\tB)}, 2.6) {Transition};
    \node[above=4pt, font=\large\bfseries, color=lineColor] at ({0.5*(\tB+\tC)}, 2.6) {Static};
    \node[above=4pt, font=\large\bfseries, color=orange!80!black] at ({0.5*(\tC+\tD)}, 2.6) {Transition};
    \node[above=4pt, font=\large\bfseries, color=lineColor] at ({0.5*(\tD+1.0)}, 2.6) {Static};


    \pgfmathsetmacro{\tauOne}{0.5*\tA}
    \coordinate (D1_P1) at (\tauOne + \TriWidth, \Ymid); 
    \coordinate (D2_P1) at (\tauOne - \TriWidth, \Ytop); 
    \coordinate (D3_P1) at (\tauOne - \TriWidth, \Ybot); 
    
    \draw[lineColor, line width=1.5pt, fill=white, fill opacity=0.6] (D1_P1) -- (D2_P1) -- (D3_P1) -- cycle;
    \fill[droneColor] (D1_P1) circle (3pt); \fill[droneColor] (D2_P1) circle (3pt); \fill[droneColor] (D3_P1) circle (3pt);
    \node[below=15pt, font=\large\bfseries, color=lineColor] at (\tauOne, 0.4) {Triangle};

    \pgfmathsetmacro{\tauTwo}{0.5*(\tB+\tC)}
    \coordinate (D2_P2) at (\tauTwo, \YlineTop); 
    \coordinate (D1_P2) at (\tauTwo, \Ymid); 
    \coordinate (D3_P2) at (\tauTwo, \YlineBot); 
    
    \draw[lineColor, line width=2pt] (D2_P2) -- (D3_P2);
    \fill[droneColor] (D1_P2) circle (3pt); \fill[droneColor] (D2_P2) circle (3pt); \fill[droneColor] (D3_P2) circle (3pt);
    \node[below=15pt, font=\large\bfseries, color=lineColor] at (\tauTwo, 0.4) {Line};

    \pgfmathsetmacro{\tauThree}{0.5*(\tD+1.0)}
    \coordinate (D1_P3) at (\tauThree + \TriWidth, \Ymid); 
    \coordinate (D2_P3) at (\tauThree - \TriWidth, \Ytop); 
    \coordinate (D3_P3) at (\tauThree - \TriWidth, \Ybot); 
    
    \draw[lineColor, line width=1.5pt, fill=white, fill opacity=0.6] (D1_P3) -- (D2_P3) -- (D3_P3) -- cycle;
    \fill[droneColor] (D1_P3) circle (3pt); \fill[droneColor] (D2_P3) circle (3pt); \fill[droneColor] (D3_P3) circle (3pt);
    \node[below=15pt, font=\large\bfseries, color=lineColor] at (\tauThree, 0.4) {Triangle};

\end{tikzpicture}}
    \caption{Visualization of the formation geometry generation over the normalized path parameter $\tau_{\text{norm}} = \tau / \tau_{\max}$. The example illustrates the smooth interpolation between defined static key-frames through dynamic transition intervals.}
    \label{fig:shape_transition}
\end{figure}
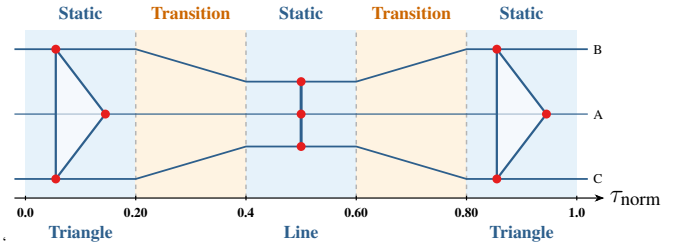

Let the continuous center trajectory be denoted by
$\mathbf p_r^c(\pathparam): [0, \pathparam_{\max}] \rightarrow \mathbb R^3.$
To allow the formation to vary continuously along the center trajectory, the desired geometry is represented by the matrix-valued function $\shapeformat(\pathparam)$, which maps each path parameter to the corresponding formation configuration.
In practice, this function is constructed from a finite set of user-defined formation key-frames. 
For any path parameter $\pathparam$ lying between two consecutive key-frames located at $\pathparam_k$ and $\pathparam_{k+1}$, the formation matrix is obtained by linear interpolation,
\begin{equation} 
\label{alg:shape_interpolation}
\mathbf M(\tau) = (1-\alpha)\mathbf M(\tau_k) + \alpha\mathbf M(\tau_{k+1}), 
\end{equation} 
where
$\alpha =
\dfrac{\pathparam-\pathparam_k}
{\pathparam_{k+1}-\pathparam_k},
\pathparam_k\le\pathparam\le\pathparam_{k+1}$.
This interpolation yields a continuous evolution of the formation geometry along the trajectory, allowing the formation to either maintain a constant geometry over an interval or transition smoothly between distinct configurations.
An example of how the formation geometry is defined by the key-frames and how it evolves with the path parameter $\tau$ is shown in \reffig{fig:shape_transition}.

The $i^{\mathrm{th}}$ column of $\shapeformat(\pathparam)$ is the relative coordinate vector
$\relativecoordvec=[x_i(\pathparam),y_i(\pathparam),z_i(\pathparam)]^T$,
representing the desired position of UAV $D_i$ with respect to the formation center. 
As the formation geometry is defined in the local formation frame, it is transformed into the world frame before generating individual UAV trajectories. 
The nominal trajectory of UAV $D_i$ in the world frame is given by
\begin{equation}
\label{eq:path_in_tau}
    \bs{p}_r^i(\tau) = \bs{p}_r^c(\tau) + \bs{R}_z\bigl(\psi(\tau)\bigr)\bs{r}_i(\tau),
\end{equation}
where $\bs{R}_z\bigl(\psi(\tau)\bigr) \in \mathbb{SO}(3)$ is the rotation matrix about the z-axis, and $\formationheading(\pathparam)$ is the formation center heading.
%

The paths $\bs{p}_r^i(\tau)$ obtained from \refeq{eq:path_in_tau} are subsequently re-parametrized by the cumulative arc length $\theta$ introduced in \refsec{subsec:formation_formulation}, as per standard \ac{MPCC} frameworks \cite{romero2022mpcc}.
Following the approach of \cite{wang2002arc}, each trajectory is sampled and approximately equidistant knot points are extracted.
These points are then used to construct a continuous, piecewise cubic spline representation of each UAV's path:

\begin{equation}\label{eq:arc_len_param_piecewise}
\bm{p}_r^i(\theta) = 
\begin{cases}
    \bm{\rho}_0(\theta), & \theta_0 \leq \theta \leq \theta_1 \\[1ex]
    \bm{\rho}_1(\theta), & \theta_1 < \theta \leq \theta_2 \\[1ex]
    \ \vdots & \quad \vdots \\[1ex]
    \bm{\rho}_{S-1}(\theta), & \theta_{S-1} < \theta \leq \theta_S
\end{cases}
\end{equation}
where $\bs{\rho}_s(\theta)$ denotes the $s$-th cubic spline segment, $\theta_0 = 0$, $\theta_S = \theta_{\max}$, and $S$ is the total number of segments used to approximate the path.

Although the generated reference trajectories $\bs{p}_r^i(\theta)$ implicitly encode the desired formation geometry, independent progress optimization causes UAVs to generally occupy different progression states, while disturbances may further induce trajectory deviations.
Therefore, the shape profile $\bs{M}(\pathparam)$ is also re-parameterized with respect to the arc-length variable $\theta$, allowing the desired formation geometry to be evaluated consistently during optimization.

For each UAV $D_i$, the arc-length re-parameterization defines a strictly monotonic forward mapping $\mathcal{F}_i:\pathparam \mapsto \theta$.
Owing to this monotonicity, there exists a well defined inverse mapping $\pathparam = \mathcal{F}^{-1}_i(\theta)$.
Substituting with this inverse mapping yields the shape profile parametrized by arc length:
\begin{equation}
\label{eq:theta_param_shape_matrix_function}
\shapeformat(\theta) := \shapeformat\left( \mathcal{F}^{-1}_i(\theta) \right).
\end{equation}
Consequently, both the reference trajectory $\bs{p}_r^i(\theta)$ and the formation geometry $\bm{M}(\theta)$ are tightly coupled and evaluated using the identical arc-length parameter $\theta$.
The next section describes how these synchronized representations are used for online formation coordination.

\subsection{In Flight Formation Coordination}
\label{subsec:formation_coordination}

In the proposed \ac{FF-MPCC}, each \ac{UAV} solves its own optimization problem independently, using the knowledge of the current and predicted positions of the remaining formation members to keep the desired formation shape.
This coordination is achieved through the formation error formulation introduced in \refsec{subsec:formation_formulation}, which requires the computation of the target position $\bs{\lambda}^i$ for ego UAV $D_i$.
This formation error is evaluated at every shooting node of the prediction horizon. 
Therefore, each UAV computes a sequence of target positions corresponding to the predicted future states considered by the optimizer. 
This requires an estimate of the predicted motion of neighbouring UAVs over the same prediction horizon.

To provide this information in a decentralized manner, every UAV periodically broadcasts both its current state and the predicted trajectory generated by its local \ac{MPCC} controller. 
At each control iteration, UAV $D_i$ combines its own predicted states with the latest predicted trajectories received from neighbouring UAVs to estimate the formation state at every prediction stage.
Using these predicted neighbour states, UAV $D_i$ constructs the measured formation position matrix $\mathbf P_m$ for each prediction stage. 
The desired formation geometry is obtained from $\shapeformat(\theta_k^i)$, where $\theta_k^i$ denotes the predicted path progression of the ego UAV at prediction stage $k$. 
A rigid-body transformation is then computed to best align the reference formation shape $\bs{M}(\theta_k^i)$ with the measured formation $\mathbf P_m$. 
Finally, the target position $\bs{\lambda}_k^i$ for the ego UAV is extracted from the transformed reference formation and used to evaluate the formation error in the optimization problem (see \refalg{alg:target_pos_computation}).

\begin{algorithm}[!htb]
\caption{Distributed Target Position Computation}
\label{alg:target_pos_computation}
\small
\DontPrintSemicolon
\KwInput{
    $\bm{P}_m \in \mathbb{R}^{3 \times N}$: Measured formation positions,
    $\shapeformat(\theta) \in \mathbb{R}^{3 \times N}$: Desired formation geometry,
    $i_{\mathrm{front}}$: Front UAV index,
    $i$: Ego UAV index
}
\KwOutput{
    $\bm{\lambda}^i \in \mathbb{R}^3$: Desired world-frame target position
    \\\vspace{0.05cm}\hrule\vspace{0.05cm}
}

$\bm{c}_m \leftarrow \frac{1}{N}\sum_{k=1}^{N}\bm{P}_m[:,k]$\tcp*{Measured center}

$\bm{c}_r \leftarrow \frac{1}{N}\sum_{k=1}^{N}\shapeformat(\theta)[:,k]$\tcp*{Reference center}

$v_{xy}\leftarrow \bm{P}_m[1\!:\!2,i_{\mathrm{front}}]-\bm{c}_m[1\!:\!2]$\tcp*{Estimate heading}

$\psi\leftarrow\mathrm{computeYaw}(v_{xy})$

$\bm{R}\leftarrow\bm{R}_z(\psi)$

$\bm{t}\leftarrow\bm{c}_m-\bm{R}\bm{c}_r$\tcp*{Rigid-body fitting}

$\bm{\lambda}^i
\leftarrow
\bm{R}\shapeformat(\theta)[:,i]+\bm{t}$\tcp*{Extract ego target}

\Return{$\bm{\lambda}^i$}

\end{algorithm}

Since the coordination involves information exchange over a wireless network, communication latency must be compensated when reconstructing the neighbour trajectories.
Let $\delta=T_0-T_s$ denote the communication delay, where $T_s$ is the message timestamp and $T_0$ is the current time. 
For prediction stage $k$, the required neighbour state corresponds to the time instant
$t_k=\delta+k\Delta T,$
where $\Delta T$ is the controller discretization timestep. 
The neighbour state is then obtained by linearly interpolating between the two predicted trajectory samples that bound $t_k$.

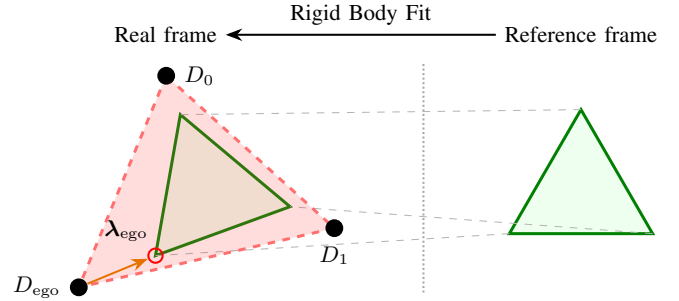
\begin{figure}[h!]
    \vspace{-1.0em}
    \centering
    \resizebox{\columnwidth}{!}{\begin{tikzpicture}[
    >=Stealth,
    drone/.style={circle, fill=black, inner sep=2.5pt},
    targetNode/.style={circle, draw=red, thick, inner sep=2.0pt},
    refShape/.style={
        draw=green!50!black,
        line width=1.2pt,
        fill=green!20,
        fill opacity=0.3
    },
    actualShape/.style={
        draw=red!70,
        dashed,
        line width=1.2pt,
        fill=red!70,
        fill opacity=0.1,
        draw opacity=0.8
    },
    vector/.style={->, thick, orange!90!black}
]

    \def\fitAngle{20}      
    \def\L{2.0}   
    
    \coordinate (FitOrigin) at (3.0,2.2);
    
    
    \coordinate (CanonOrigin) at (8.2,2.2);

    \pgfmathsetmacro{\Height}{sqrt(3)/2*\L}
    
    \coordinate (A0) at (0,  2*\Height/3);
    \coordinate (B0) at (-\L/2, -\Height/3);
    \coordinate (C0) at (\L/2, -\Height/3);
    
    \coordinate (LA) at ($(CanonOrigin)+(A0)$);
    \coordinate (LB) at ($(CanonOrigin)+(B0)$);
    \coordinate (LC) at ($(CanonOrigin)+(C0)$);
    
    \draw[refShape] (LA)--(LB)--(LC)--cycle;
    
    
    
    \coordinate (RA) at ($(FitOrigin)+({cos(\fitAngle)*0 - sin(\fitAngle)*(2*\Height/3)},
                                       {sin(\fitAngle)*0 + cos(\fitAngle)*(2*\Height/3)})$);
    
    \coordinate (RB) at ($(FitOrigin)+({cos(\fitAngle)*(-\L/2) - sin(\fitAngle)*(-\Height/3)},
                                       {sin(\fitAngle)*(-\L/2) + cos(\fitAngle)*(-\Height/3)})$);
    
    \coordinate (RC) at ($(FitOrigin)+({cos(\fitAngle)*(\L/2) - sin(\fitAngle)*(-\Height/3)},
                                       {sin(\fitAngle)*(\L/2) + cos(\fitAngle)*(-\Height/3)})$);

    \draw[refShape] (RA)--(RB)--(RC)--cycle;
    
    
    \draw[-,gray!60,dashed] (LA) -- (RA);
    \draw[-,gray!60,dashed] (LB) -- (RB);
    \draw[-,gray!60,dashed] (LC) -- (RC);
    
    
    \def\actualScale{1.5}    
    
    \def\Ax{0.0}
    \def\Ay{0.0}
    
    \def\Bx{-0.7}
    \def\By{0.0}
    
    \def\Cx{0.05}
    \def\Cy{-0.2}
    
    \coordinate (PA) at ($ (FitOrigin)!\actualScale!(RA) + (\Ax,\Ay) $);
    \coordinate (PB) at ($ (FitOrigin)!\actualScale!(RB) + (\Bx,\By) $);
    \coordinate (PC) at ($ (FitOrigin)!\actualScale!(RC) + (\Cx,\Cy) $);
    
    \fill[red!70, fill opacity=0.10]
        (PA) -- (PB) -- (PC) -- cycle;
    
    \draw[actualShape]
        (PA) -- (PB) -- (PC) -- cycle;
    
    \node[drone, label={left:\small $D_{\mathrm{ego}}$}] (pB) at (PB) {};
    \node[drone, label={below:\small $D_1$}]  (pC) at (PC) {};
    \node[drone, label={right:\small $D_0$}] (pA) at (PA) {};

    \node[targetNode,
          label={[xshift=1mm]above left:\small $\boldsymbol{\lambda}_{\mathrm{ego}}$}]
          (targetB) at (RB) {};

    \draw[vector] (pB) -- (targetB);

    
    \draw[densely dotted, thick, gray!70]
        (6.0,0.8) -- (6.0,4.0);
    
    \node[font=\small] (realFrame) at (2.4,4.4) {Real frame};
    \node[font=\small] (refFrame)  at (8.2,4.4) {Reference frame};
    
    \draw[->, thick]
        (refFrame.west) -- (realFrame.east)
        node[midway, above] {\small Rigid Body Fit};
    
\end{tikzpicture}}
    \caption{
    Distributed rigid-body fitting. The green and red dashed triangles denote the desired fitted and the current formation, respectively. The target position $\bs{\lambda}_{\mathrm{ego}}$ is computed from the estimated positions of $D_0$ and $D_1$. Each UAV performs the procedure independently using local information.
    \vspace{-1.0em}
    }
    \label{fig:formulation_fitting_tikz}
\end{figure}

The proposed distributed approach eliminates the need for centralized coordination and allows each UAV to continuously determine its desired position within the formation based on its predicted state and the predicted neighbour states received through communication. 
In contrast to other methods, our proposed method allows the formation to  adapt to in-flight disturbances and also to slower formation members, while maintaining the desired geometric structure. 
A simple illustration of the fitting process is shown in \reffig{fig:formulation_fitting_tikz}.

\section{Results}

The performance of the proposed \ac{FF-MPCC} is demonstrated in extensive simulations and  real world flights.
Simulations and real word experiments are conducted with identical custom agile quadrotor model measuring \qty{300}{\milli\meter} in diagonal length and weighing \qty{1.2}{\kilogram}. 
Each quadcopter is equipped with RTK GPS, a CubePilot flight controller with PX4 firmware and is commanded by the MRS UAV system architecture \cite{mrs_uav_sustem} running on Khadas Vim3 Pro single-board computer.
The collective thrust of the UAVs is capped at \qty{48}{\newton} in the controller constraints, while the per motor thrust is capped at \qty{17}{\newton}.
The position data of the neighboring robots in the formation is shared over a standard WiFi interface at \qty{10}{\hertz} frequency with an average delay of \qty{15}{\milli \second}. 

\subsection{Experiment Methodology}
\label{subsec:experimentr_methodology}

The goal of \ac{FF-MPCC} is to enable fast formation flight; therefore, we rely on formation accuracy and flight completion time to evaluate its performance.
The formation flight accuracy is evaluated using two metrics, both computed over uniformly sampled trajectory measurements collected during the experiment.
The first is the \acf{MAPDE}:
\begin{equation}
\label{eq:mapde_formula}
\mathrm{MAPDE} = \frac{1}{T \binom{N}{2}} \sum_{k=1}^{T} \sum_{i=1}^{N-1} \sum_{j=i+1}^{N} \left| d_{ij}(k) - d_{ij}^{\ast}(k) \right|,
\end{equation}
where $\left| d_{ij}(k)-d_{ij}^{\ast}(k) \right|$ is the pairwise distance error for the UAV pair $\{D_i, D_j\}$. Here, $d_{ij}(k)$ and $d_{ij}^{\ast}(k)$ represent the actual and desired mutual Euclidean distances at time step $k$, respectively, $T$ denotes the total number of recorded trajectory samples, and $N$ is the number of UAVs. 
The desired pairwise distances are computed based on the reference shape matrix $\bs{M}(\theta_k^i)$ \refeq{eq:theta_param_shape_matrix_function}, where $\theta_k^i$ denotes the progress of $D_i$ at timestep $k$.
Since neighboring UAVs may evaluate the desired geometry at slightly different progression values, the pairwise reference distances are computed independently for each UAV.

While \ac{MAPDE} evaluates relative spacing, it fails to capture the overall orientation (heading) of the formation and does not scale reliably for larger formations ($N > 3$). 
To address these limitations, we introduce the \ac{SFTE}. 
The \ac{SFTE} at any given time step $k$ is computed following these steps:
\begin{enumerate}
    \item \textbf{Orientation Alignment:} The actual center of formation $\bm{p}^c(k)$ is computed and projected onto the reference center path $\bs{p}_r^c$ (defined in \refsec{subsec:formation_path_generation}) to locate the closest Euclidean point. The path tangent at this point yields the desired formation heading $\psi^c(k)$, which defines the rotation matrix $\bm{R}(\psi)$ used to orient $\bs{M}(\theta_k^i)$.
    \item \textbf{Anchoring:} To evaluate shape distortion independently of global translation errors, each UAV $D_i$ is sequentially treated as a local coordinate anchor. The rotated shape matrix is translated such that the expected position of $D_i$ aligns with its actual recorded position.
    \item \textbf{Error Evaluation:} With the shape matrix rotated by $\psi^c(k)$ and anchored at $D_i$, the Euclidean distance error between the actual position of every other UAV $D_j$ ($j \neq i$) and its corresponding target position $\bm{\lambda}^j_{i, k}$ on the matrix is computed. 
\end{enumerate}
Formally, the \ac{SFTE} is given by
\begin{equation}
\label{eq:sfte_formula}
\mathrm{SFTE} = \frac{1}{T N (N-1)} \sum_{k=1}^{T} \sum_{i=1}^{N} \sum_{\substack{j=1 \\ j \neq i}}^{N} \left\| \bm{p}^j_k - \bs{\lambda}^j_{i, k} \right\|,
\end{equation}
where $\bs{p}^j_k$ is the actual position of UAV $D_j$ at timestep $k$.

Mission completion time is defined as the time required for the entire formation to reach its target destination. 
A \ac{UAV} is considered to have reached its goal when it enters a spherical region of radius \SI{0.75}{\meter} centered at its assigned endpoint
Thus, the mission completion time corresponds to the maximum arrival time among all \acp{UAV}.
This definition is used consistently across all experiments.

\subsection{Simulation Results}
\label{subsec:simulation_results}

We evaluate our method through Gazebo simulations. We first present an ablation study to isolate the impact of our formation error weight, followed by a comparative evaluation against NMPC-based approach.

\subsubsection{Ablation Study}
\label{subsubsec:ablation_study}

We conducted an ablation study to demonstrate the necessity of the formation error cost introduced in \refeq{eq:formation_error_basic}. 
The evaluation scenario tasks a team of three UAVs ($N=3$) with tracking a sine wave center path while maintaining a static equilateral triangular shape.
To establish the ablation baseline, we first run the simulation with the formation error weight $q_f$ set to 0 in the objective function \refeq{eq:ff_mpcc_ocp_objective}, eliminating the influence of the formation error in the cost, thus, causing the UAVs to track their trajectories independently.
We then rerun the exact same path using a sweep of different non-zero values for $q_f$ to analyze the effects on all metrics.
\begin{figure}[!htbp]
    \vspace{-1em}
    \centering
    \begin{subfigure}{0.49\linewidth}
        \centering
        \includegraphics[width=\linewidth]{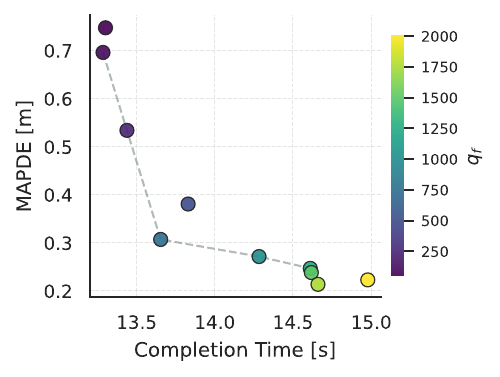}
        \phantomsubcaption\label{subfig:mapde_ablation}
    \end{subfigure}
    \hfill
    \begin{subfigure}{0.49\linewidth}
        \centering
        \includegraphics[width=\linewidth]{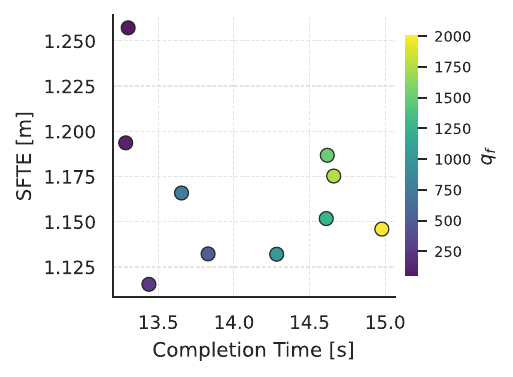}
        \phantomsubcaption\label{subfig:sfte_ablation}
    \end{subfigure}
    \vspace{-2.8em}
    \caption{Ablation study of the influence of configuration parameter $q_f$ on formation completion time and formation accuracy metrics evaluated using 3 agents scenarios. Performance profiles are color-coded by parameter value.
    }
    \label{fig:dist_gains_ablation}
\end{figure}

The results of the ablation study are summarized in \reffig{fig:dist_gains_ablation}.
The baseline with $q_f = 0$, which corresponds to the traditional \ac{MPCC}, is omitted from the plots as it is a clear outlier with a \ac{MAPDE} of \SI{1.82}{\meter} and an \ac{SFTE} of \SI{3.41}{\meter}, at a completion time of \SI{12.55}{\second}.
In contrast, all non-zero values of $q_f$ result in \ac{MAPDE} within \qtyrange{0.25}{0.78}{\meter} and \ac{SFTE} within \qtyrange{1.08}{1.3}{\meter}, at completion times within \qtyrange{13.21}{14.98}{\second}.
The introduced formation error thus improves the formation accuracy by at least $57\%$ in \ac{MAPDE} and $62\%$ in \ac{SFTE}.
The results in \reffig{fig:dist_gains_ablation}\subref{subfig:mapde_ablation} show that increasing $q_f$ creates a clear trade-off: higher penalty weights systematically drive down the \ac{MAPDE}, but they increase the total completion time.
This is a result of slowdown in individual UAVs to maintain cohesion in challenging sections of the trajectory, causing a slowdown overall.
The \ac{SFTE} in \reffig{fig:dist_gains_ablation}\subref{subfig:sfte_ablation} does not follow the same trend as it is largely influenced by rigid-body orientation errors rather than exact distances.
However, it stays consistently low for all non-zero $q_f$, which indicates that even small $q_f$ values are enough to maintain the formation orientation.

\subsubsection{Comparative Evaluation}
\label{subsubsec:comparitive_nmpc_eval}

\begin{table*}[!ht]
\centering
\caption{Comparison of the baseline PMM+NMPC and the proposed \ac{FF-MPCC} in formation flight scenarios. All simualtions are conducted with identical constant gains, and the reported results are averaged over 20 trials per scenario.}
\vspace{-0.5em}
\label{tab:comparison_with_nmpc}
\begin{tabular}{llcccccccc}
\toprule
\multirow{2}{*}[-0.9ex]{Path type} &
\multirow{2}{*}[-0.9ex]{Formation} &
\multirow{2}{*}[-0.9ex]{Alignment} &
\multirow{2}{*}[-0.7ex]{\shortstack{Path\\length [m]}} &
\multicolumn{3}{c}{PMM+NMPC} &
\multicolumn{3}{c}{FF-MPCC} \\
\cmidrule(lr){5-7}
\cmidrule(lr){8-10}
&
&
&
&
\ac{SFTE} [m] &
\ac{MAPDE} [m] &
Time [s] &
\ac{SFTE} [m] &
\ac{MAPDE} [m] &
Time [s] \\
\midrule
Straight Line &
Triangle &
Yes &
60 &
0.095 & 0.056 & \textbf{4.406} &
\textbf{0.093} & \textbf{0.049} & 5.341 \\

Straight Line &
Triangle-Line &
Yes & 
60 &
0.935 & 0.532 & \textbf{4.443} &
\textbf{0.250} & \textbf{0.176} & 5.084\\

Straight Line &
Triangle-Line-Triangle &
Yes &
60 &
1.764 & 1.051 & \textbf{4.927} &
\textbf{1.195} & \textbf{0.397} & 6.361 \\

Spiral &
Triangle &
Yes &
62.93 &
2.643 & 0.116 & 6.139 &
\textbf{0.262} & \textbf{0.081} & \textbf{5.963}\\

Spiral &
Triangle-Line &
Yes &
62.93 &
3.371 & 0.305 & \textbf{5.910} & 
\textbf{0.223} & \textbf{0.107} & 6.285\\

Spiral &
Triangle-Line-Triangle &
Yes &
62.93 &
3.567 & 0.388 & \textbf{6.094} & 
\textbf{0.466} & \textbf{0.183} & 7.324\\

Sine Wave &
Triangle & 
Yes &
195.19 &
2.985 & 0.662 & 16.897 &
\textbf{1.186} & \textbf{0.238} & \textbf{14.618}\\

Sine Wave &
Triangle-Line & 
Yes &
195.19 &
2.267 & 0.724 & 16.456 &
\textbf{1.006} & \textbf{0.353} & \textbf{13.691}\\

Sine Wave &
Triangle-Line-Triangle & 
Yes &
195.19 &
3.239 & 0.932 & 16.974 &
\textbf{0.992} & \textbf{0.323} & \textbf{13.535}\\

\bottomrule
\vspace{-2.7em}
\end{tabular}
\end{table*}

To evaluate the overall performance of our proposed framework, we compare it against a conventional planning-and-control pipeline, further denoted as PMM+NMPC, that combines minimum-time trajectory planning with \ac{NMPC}-based trajectory tracking.
The setup of PMM+NMPC involves applying PMM~\cite{teissing2024PMM} to generate time optimal trajectory of the defined formation center $p_c^d$.
The center trajectory is then offset using the formation geometry defined in \refeq{eq:theta_param_shape_matrix_function} to generate individual UAV trajectories for tracking. 
To compute \ac{MAPDE} we use the reference pairwise distances obtained from the shape matrix at time $k$. 
For \ac{SFTE} and completion time, we employ the same method described in \refsec{subsec:experimentr_methodology}, ensuring evaluation consistency. 

\begin{figure}[htbp]
    \vspace{-0.1cm}
    \centering
    \begin{subfigure}{\linewidth}
        \centering
        \includegraphics[width=\linewidth]{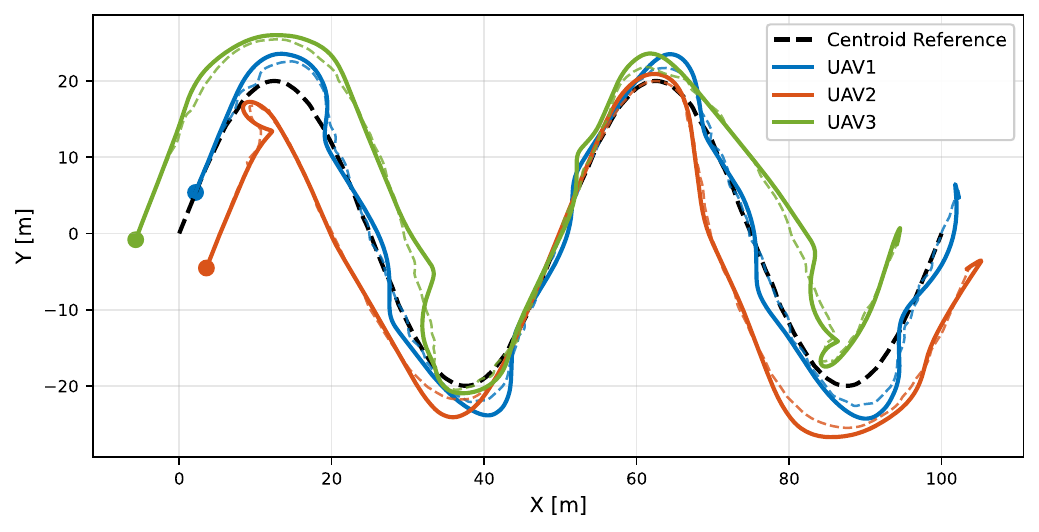}\vspace{-0.3cm}
        \caption{PMM+NMPC trajectory Profile}
        \label{subfig:nmpc_plot}
    \end{subfigure}
    
    \vspace{-0.0cm} 
    
    \begin{subfigure}{\linewidth}
        \centering
        \includegraphics[width=\linewidth]{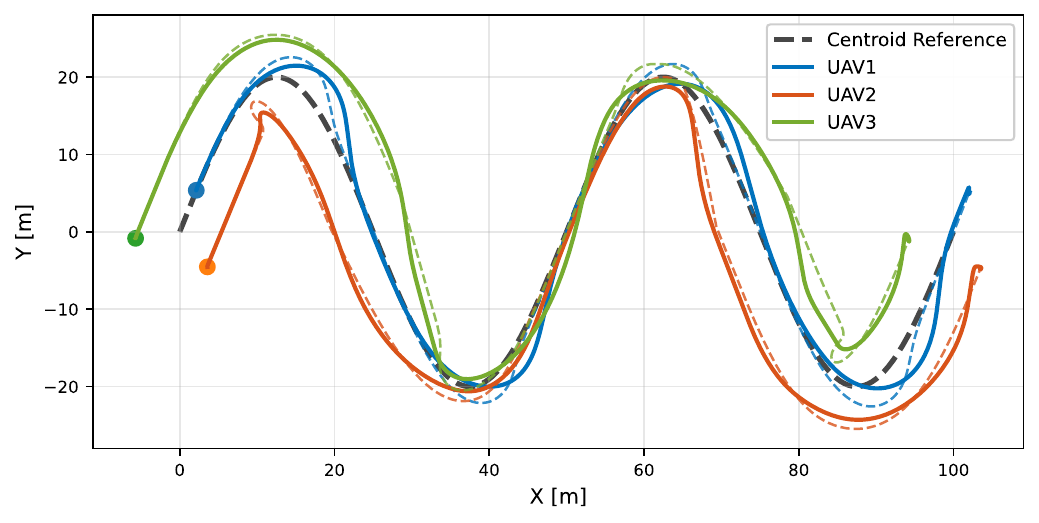}\vspace{-0.3cm} 
        \caption{FF-MPCC trajectory Profile}
        \label{subfig:comparison_plot}
    \end{subfigure}\vspace{-0.1cm}

    \caption{Trajectory comparison between the baseline PMM+NMPC and FF-MPCC for a sine-wave path with a triangle-line-triangle formation geometry. 
    Dashed lines denote the reference trajectories generated by each method, solid lines the trajectories flown by the individual UAVs.
    }
    \label{fig:trajectory_comparison}
    \vspace{-2.0em}
\end{figure}

\begin{figure}[t]
\vspace{-0.7em}
    \centering
    \includegraphics[width=0.83\columnwidth]{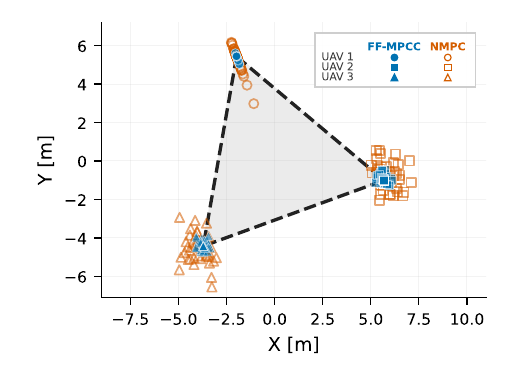}
    \vspace{-1.0em}
    \caption{Scatter of actual UAV positions around the reference shape after transformation to the reference frame for the Sine Wave with static Triangle formation listed in Table~\ref{tab:comparison_with_nmpc}. The dashed triangle denotes the desired formation. The formation is centered at its centroid, with UAV~1 defining the orientation of the reference frame.}
    \label{fig:formation_comparison_scatter}
    \vspace{-1.4em}
\end{figure}

\ac{FF-MPCC} consistently outperforms PMM+NMPC in both \ac{MAPDE} and \ac{SFTE} in all simulations with margins greater than 30\% while also providing faster completion times in half the tested scenarios (see \reftab{tab:comparison_with_nmpc} and \reffig{fig:formation_comparison_scatter}).
While \ac{FF-MPCC} exhibits longer completion times in some cases, for every scenario there exist gain settings that match the completion times of PMM+NMPC while still achieving superior formation accuracy.
The advantages of the proposed method are more pronounced in challenging scenarios with curved trajectories and complex dynamic formation timelines. 
\ac{FF-MPCC} can adapt to non feasible paths with greater agility while also maintaining formation integrity at sharp curves. 
The trajectory plots of a sample scenario for both the methods, highlighting these effects, can be seen in \reffig{fig:trajectory_comparison}.

\subsection{Real World Experiment}
\label{subsec: real_world_results}
\begin{figure*}[!ht]
    \centering

    \begin{subfigure}[t]{0.48\linewidth}
        \centering
        \includegraphics[width=\linewidth]{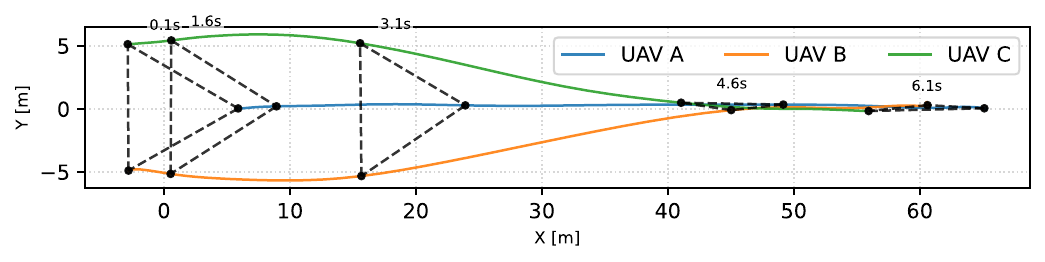}\\[0.3em]
        \includegraphics[width=\linewidth,trim={0 4cm 0 4cm},clip]{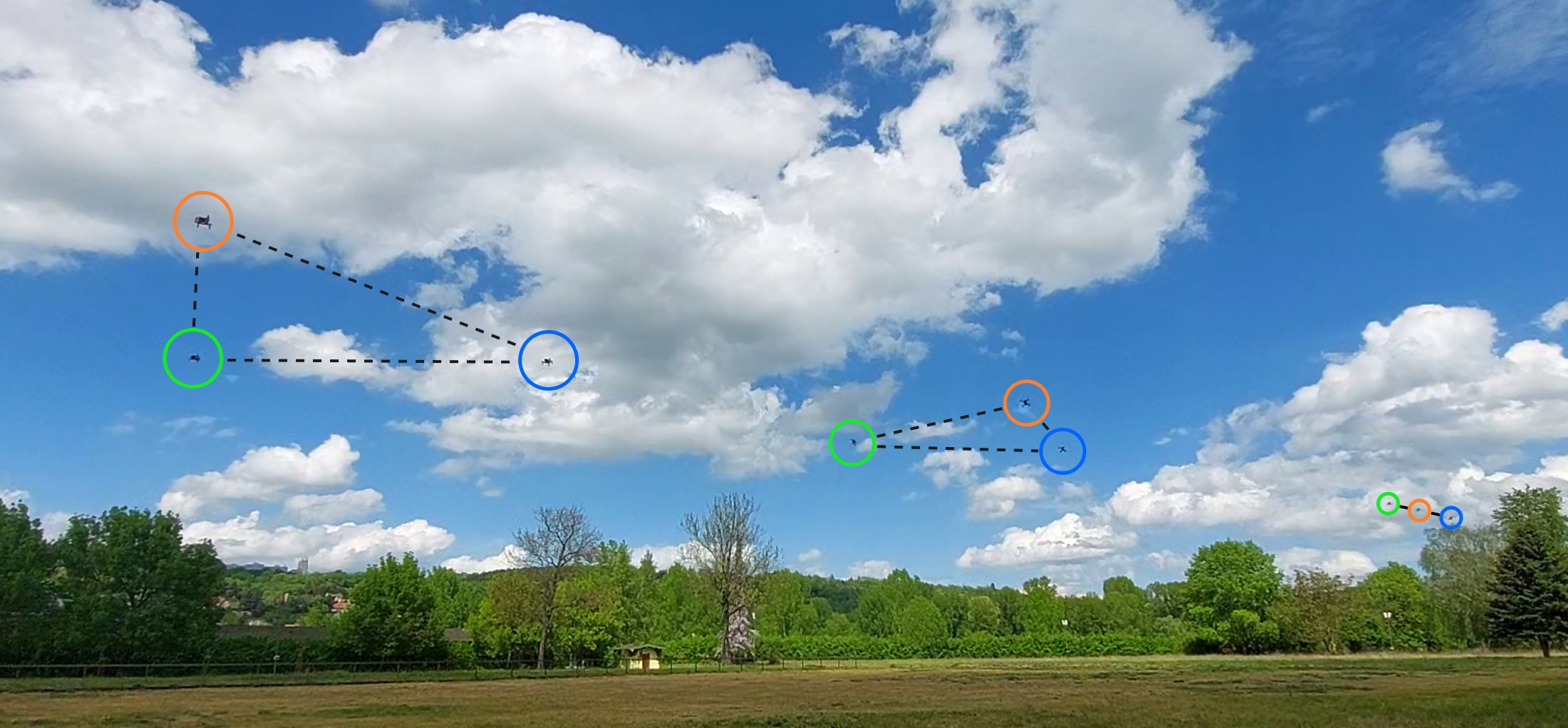}\vspace{-0.4em}
        \caption{Triangle-Line transition}
        \label{fig:scenario_a}
    \end{subfigure}\hfill%
    \begin{subfigure}[t]{0.48\linewidth}
        \centering
        \includegraphics[width=\linewidth]{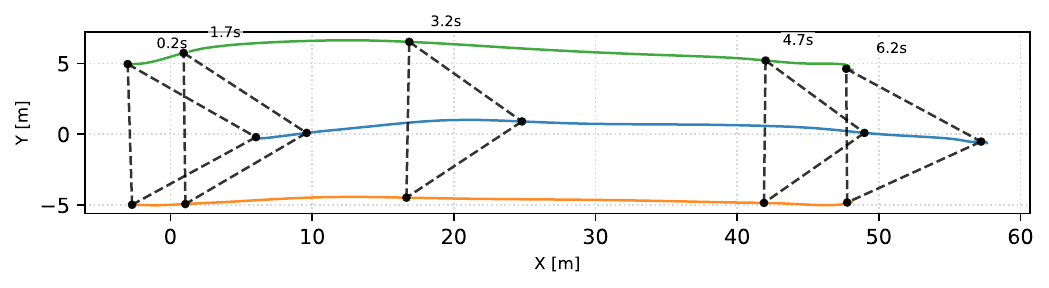}\\[0.3em]
        \includegraphics[width=\linewidth,trim={0 4cm 0 4cm},clip]{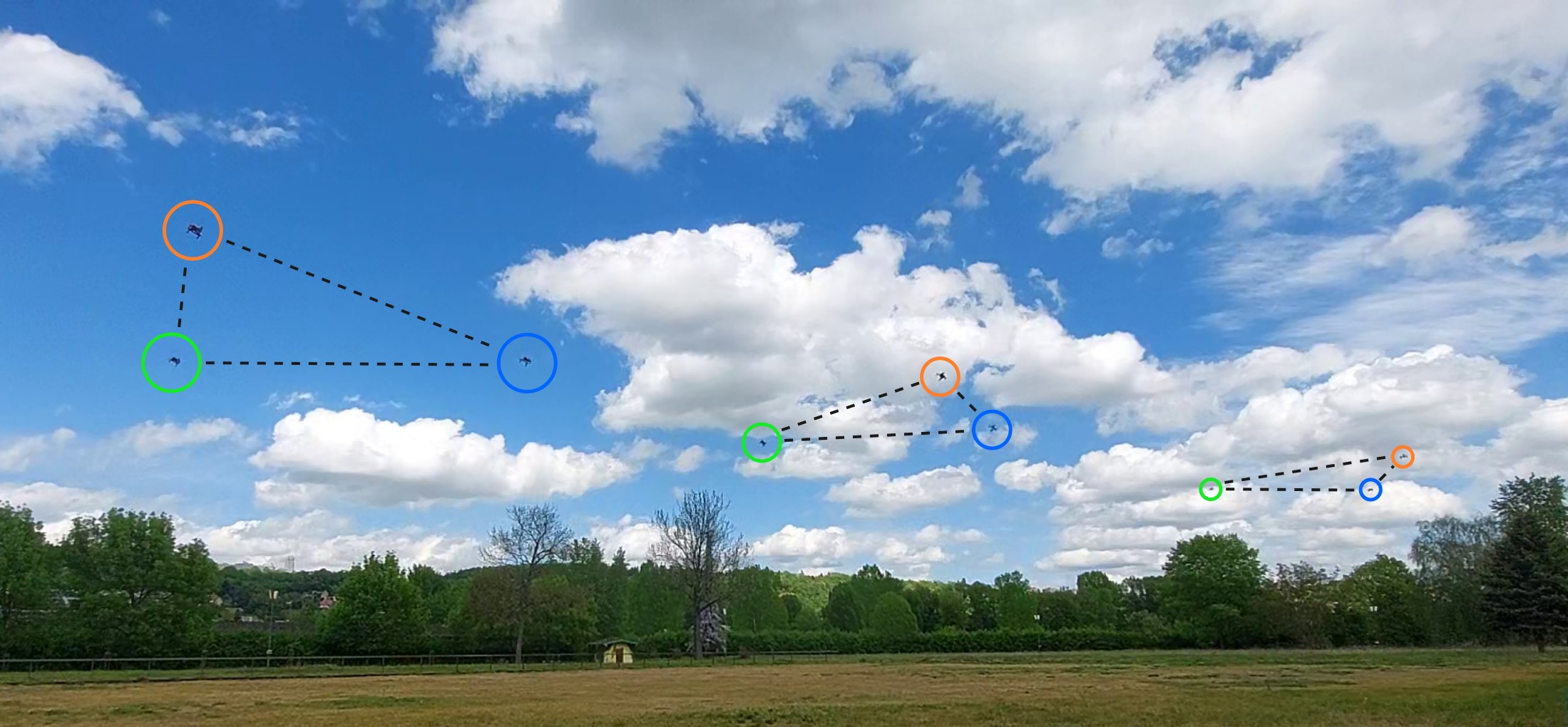}\vspace{-0.4em}
        \caption{Triangle only}
        \label{fig:scenario_b}
    \end{subfigure}
    \vspace{-0.5em}
    \caption{Real-world validation experiments. The top image in each column shows the reconstructed trajectories, while the bottom image shows the corresponding captured experimental scene. The global formation path for both scenarios is a straight line, where a) depicts a transitional geometry from triangle to straight line while b) shows a static triangular geometry.}
    \label{fig:real_world_validation}
    \vspace{-1.5em}
\end{figure*}

Field experiments were conducted using a three-\ac{UAV} scenario to validate the feasibility of the proposed decentralized \ac{FF-MPCC} framework under real-world conditions. 
The experimental flights were performed along a \SI{60}{\meter} straight-line trajectory and evaluated two representative formation profiles.
The first scenario maintained a static equilateral triangle with a side length of \SI{10}{\meter}. 
The second scenario transitioned from the same triangular configuration to a co-linear formation with a total length of \SI{8.66}{\meter}. 
The successful execution of both scenarios demonstrates the ability of the proposed framework to maintain and reconfigure formation geometry during flight without centralized coordination under real-world conditions.
The reconstructed trajectories and corresponding images for both scenarios are shown in \reffig{fig:real_world_validation}.

\section{Conclusion}

This paper presents \ac{FF-MPCC}, a distributed \ac{MPCC}-based framework for decentralized multi-\ac{UAV} formation flight. The proposed approach jointly optimizes path progression and formation accuracy, enabling agile and time-efficient formation flight along predefined paths without requiring offline generation of dynamically feasible trajectories. By combining a heading-aware formation representation with distributed rigid-body fitting, each \ac{UAV} independently optimizes its motion while maintaining the prescribed, dynamically evolving formation geometry. The presented results demonstrate that the advantages of \ac{MPCC}, previously established for single-\ac{UAV} flight, can be successfully extended to fast decentralized formation flight, providing a practical solution for agile multi-\ac{UAV} operations under realistic communication conditions.

\bibliographystyle{IEEEtran}
\bibliography{main.bib}

@article{kratky2025catora,
  author   = {Kratky, Vit and Penicka, Robert and Horyna, Jiri and Stibinger, Petr and Baca, Tomas and Petrlik, Matej and Stepan, Petr and Saska, Martin},
  journal  = {IEEE Transactions on Robotics},
  pages    = {2950-2969},
  title    = {{CAT-ORA: Collision-Aware Time-Optimal Formation Reshaping for Efficient Robot Coordination in 3-D Environments}},
  volume   = 41,
  year     = 2025,
  doi      = {10.1109/TRO.2025.3547296},
}

@article{romero2022mpcc,
  title={{Model predictive contouring control for time-optimal quadrotor flight}},
  author={Romero, Angel and Sun, Sihao and Foehn, Philipp and Scaramuzza, Davide},
  journal={IEEE Transactions on Robotics},
  volume={38},
  number={6},
  pages={3340--3356},
  year={2022},
  publisher={IEEE}
}

@article{romero2022replanning,
  title    = {{Time-optimal online replanning for agile quadrotor flight}},
  author   = {Romero, Angel and Penicka, Robert and Scaramuzza, Davide},
  journal  = {IEEE Robotics and Automation Letters},
  volume   = {7},
  number   = {3},
  pages    = {7730--7737},
  year     = {2022},
}

@Inbook{deihl2007shootingnodes,
author="Diehl, M.
and Bock, H.G.
and Diedam, H.
and Wieber, P.-B.",
editor="Diehl, Moritz
and Mombaur, Katja",
title={{Fast Direct Multiple Shooting Algorithms for Optimal Robot Control}},
bookTitle="Fast Motions in Biomechanics and Robotics: Optimization and Feedback Control",
year="2006",
publisher="Springer Berlin Heidelberg",
pages="65--93",
isbn="978-3-540-36119-0",
doi="10.1007/978-3-540-36119-0_4",
}

@misc{ji2021cmpcccorridorbasedmodelpredictive,
  title         = {{CMPCC: Corridor-based Model Predictive Contouring Control for Aggressive Drone Flight}},
  author        = {Jialin Ji and Xin Zhou and Chao Xu and Fei Gao},
  year          = {2021},
  eprint        = {2007.03271},
  archivePrefix = {arXiv},
  primaryClass  = {cs.RO},
  note          = {arXiv:2007.03271 [cs.RO]},
  url           = {https://arxiv.org/abs/2007.03271}
}

@misc{guevara2024modelpredictivecontouringcontrol,
  title        = {{Model Predictive Contouring Control with Barrier and Lyapunov Functions for Stable Path-Following in UAV systems}},
  author       = {Bryan S. Guevara and Viviana Moya and Luis F. Recalde and David Pozo-Espin and Daniel C. Gandolfo and Juan M. Toibero},
  year         = {2024},
  eprint       = {2411.00668},
  archivePrefix = {arXiv},
  primaryClass = {cs.RO},
  note         = {arXiv:2411.00668 [cs.RO]},
  url          = {https://arxiv.org/abs/2411.00668}
}

@article{Liniger_2014,
   title={{Optimization‐Based Autonomous Racing of 1:43 Scale RC Cars}},
   volume={36},
   ISSN={1099-1514},
   DOI={10.1002/oca.2123},
   number={5},
   journal={Optimal Control Applications and Methods},
   publisher={Wiley},
   author={Liniger, Alexander and Domahidi, Alexander and Morari, Manfred},
   year={2014}, 
   pages={628-647}
   }

@article{wang2021uav,
  title={{UAV formation obstacle avoidance control algorithm based on improved artificial potential field and consensus}},
  author={Wang, Ning and Dai, Jiyang and Ying, Jin},
  journal={International Journal of Aeronautical and Space Sciences},
  volume={22},
  number={6},
  pages={1413--1427},
  year={2021},
  publisher={Springer}
}

@article{seo2017collision,
  title={{Collision avoidance strategies for unmanned aerial vehicles in formation flight}},
  author={Seo, Joongbo and Kim, Youdan and Kim, Seungkeun and Tsourdos, Antonios},
  journal={IEEE Transactions on aerospace and electronic systems},
  volume={53},
  number={6},
  pages={2718--2734},
  year={2017},
  publisher={IEEE}
}

@article{saska2020formation,
  title={{Formation control of unmanned micro aerial vehicles for straitened environments}},
  author={Saska, Martin and Hert, Daniel and Baca, Tomas and Kratky, Vit and Nascimento, Tiago},
  journal={Autonomous Robots},
  volume={44},
  number={6},
  pages={991--1008},
  year={2020},
  publisher={Springer},
  doi={10.1007/s10514-020-09913-0}
}

@INPROCEEDINGS{nguyen2021mpcsurvey,
  author={Nguyen, Huan and Kamel, Mina and Alexis, Kostas and Siegwart, Roland},
  booktitle={2021 European Control Conference (ECC)}, 
  title={{Model Predictive Control for Micro Aerial Vehicles: A Survey}}, 
  year={2021},
  pages={1556-1563},
  doi={10.23919/ECC54610.2021.9654841}
  }

@article{penicka2022mintimeplanning,
  title    = {{Minimum-time quadrotor waypoint flight in cluttered environments}},
  author   = {Penicka, Robert and Scaramuzza, Davide},
  journal  = {IEEE Robotics and Automation Letters},
  volume   = {7},
  number   = {2},
  pages    = {5719--5726},
  year     = {2022},
  doi      = {10.1109/LRA.2022.3154013},
}

@ARTICLE{teissing2024PMM,
  author={Teissing, Krystof and Novosad, Matej and Penicka, Robert and Saska, Martin},
  journal={IEEE Robotics and Automation Letters}, 
  title={{Real-Time Planning of Minimum-Time Trajectories for Agile UAV Flight}}, 
  year={2024},
  volume={9},
  number={11},
  pages={10351-10358},
}

@INPROCEEDINGS{gupta2025lolNmpc,
  author={Gupta, Parakh M. and Procházka, Ondřej and Hřebec, Jan and Novosad, Matej and Pěnička, Robert and Saska, Martin},
  booktitle={2025 IEEE/RSJ International Conference on Intelligent Robots and Systems (IROS)}, 
  title={{LoL-NMPC: Low-Level Dynamics Integration in Nonlinear Model Predictive Control for Unmanned Aerial Vehicles}}, 
  year={2025},
  pages={1186-1193},
}

@inproceedings{wang2002arc,
  title={{Arc-length parameterized spline curves for real-time simulation}},
  author={Wang, Hongling and Kearney, Joseph and Atkinson, Kendall},
  booktitle={Proc. 5th international conference on curves and surfaces},
  volume={387396},
  year={2002}
}

@ARTICLE{quan2023formationFlightInDenseEnvironments,
  author={Quan, Lun and Yin, Longji and Zhang, Tingrui and Wang, Mingyang and Wang, Ruilin and Zhong, Sheng and Zhou, Xin and Cao, Yanjun and Xu, Chao and Gao, Fei},
  journal={IEEE Transactions on Robotics}, 
  title={{Robust and Efficient Trajectory Planning for Formation Flight in Dense Environments}}, 
  year={2023},
  volume={39},
  number={6},
  pages={4785-4804},
  doi={10.1109/TRO.2023.3301295}}

@ARTICLE{deng2026mpccFormation,
  author={Deng, Jian and Deng, Yimin and Duan, Haibin},
  journal={IEEE Transactions on Circuits and Systems I: Regular Papers}, 
  title={{Bioinspired Homotopic Model Predictive Contouring Control for Fixed-Wing Unmanned Aerial Vehicle Swarm}}, 
  year={2026},
  volume={73},
  number={3},
  pages={2143-2155},
  doi={10.1109/TCSI.2025.3598794}}

@article{sun2022comparative,
  title={{A comparative study of nonlinear mpc and differential-flatness-based control for quadrotor agile flight}},
  author={Sun, Sihao and Romero, Angel and Foehn, Philipp and Kaufmann, Elia and Scaramuzza, Davide},
  journal={IEEE Transactions on Robotics},
  volume={38},
  number={6},
  pages={3357--3373},
  year={2022},
  publisher={IEEE}
}

@article{gomaa2022computationally,
  title={{Computationally efficient stability-based nonlinear model predictive control design for quadrotor aerial vehicles}},
  author={Gomaa, Mahmoud AK and De Silva, Oscar and Mann, George KI and Gosine, Raymond G},
  journal={IEEE Transactions on Control Systems Technology},
  volume={31},
  number={2},
  pages={615--630},
  year={2022},
  publisher={IEEE}
}

@ARTICLE{pan2025tstar,
  author={Pan, Honghao and Zahmatkesh, Mohsen and Rekabi-Bana, Fatemeh and Arvin, Farshad and Hu, Junyan},
  journal={IEEE Transactions on Intelligent Transportation Systems}, 
  title={{T-STAR: Time-Optimal Swarm Trajectory Planning for Quadrotor Unmanned Aerial Vehicles}}, 
  year={2025},
  volume={26},
  number={8},
  pages={12532-12547},
  doi={10.1109/TITS.2025.3557783}}

@INPROCEEDINGS{guan2025learningDmpccWithOrca,
  author={Guan, Xin and Zhao, Fangguo and Tian, Shunxin and Li, Shuo},
  booktitle={2025 IEEE International Conference on Robotics and Automation (ICRA)}, 
  title={{Learning Time-Optimal Online Replanning for Distributed Model Predictive Contouring Control of Quadrotors}}, 
  year={2025},
  volume={},
  number={},
  pages={14527-14533},
  doi={10.1109/ICRA55743.2025.11128315}}

@INPROCEEDINGS{brust2015envmapping,
  author={Brust, Matthias R. and Strimbu, Bogdan M.},
  booktitle={2015 IEEE Tenth International Conference on Intelligent Sensors, Sensor Networks and Information Processing (ISSNIP)}, 
  title={{A networked swarm model for UAV deployment in the assessment of forest environments}}, 
  year={2015},
  volume={},
  number={},
  pages={1-6},
  doi={10.1109/ISSNIP.2015.7106967}}

@ARTICLE{petracek2024historicpreservation,
  author={Petracek, Pavel and Kratky, Vit and Baca, Tomas and Petrlik, Matej and Saska, Martin},
  journal={IEEE Robotics \& Automation Magazine}, 
  title={{New Era in Cultural Heritage Preservation: Cooperative Aerial Autonomy for Fast Digitalization of Difficult-to-Access Interiors of Historical Monuments}}, 
  year={2024},
  volume={31},
  number={2},
  pages={8-25},
  doi={10.1109/MRA.2023.3244423}}

@article{ccari2023novelformationtransport,
  title={{A novel neural network-based robust adaptive formation control for cooperative transport of a payload using two underactuated quadcopters}},
  author={Ccari, Luis F Canaza and Yanyachi, Pablo Raul},
  journal={IEEE Access},
  volume={11},
  pages={36015--36028},
  year={2023},
  publisher={IEEE}
}

@ARTICLE{zhang2025agile,
  author={Zhang, Jingsen and Hou, Biao and Huang, Rui},
  journal={IEEE Robotics and Automation Letters}, 
  title={{Agile Trajectory Planning and Large Obstacle Avoidance for Formation Flight Using a Virtual Core}}, 
  year={2025},
  volume={10},
  number={8},
  pages={8546-8553},
  doi={10.1109/LRA.2025.3586516}}

@ARTICLE{zhou2025reformation,
  author={Zhou, Yuan and Quan, Lun and Xu, Chao and Xu, Guangtong and Gao, Fei},
  journal={IEEE Transactions on Automation Science and Engineering}, 
  title={{RE-Formation: Resilient and Efficient Formation Planning in Large-Scale Distributed Aerial Swarms}}, 
  year={2025},
  volume={22},
  number={},
  pages={21212-21227},
  doi={10.1109/TASE.2025.3603614}}

@ARTICLE{mrs_uav_sustem,
author = {Báča, Tomáš and Petrlík, Matěj and Vrba, Matouš and Spurný, Vojtěch and Pěnička, Robert and Hert, Daniel and Saska, Martin},
year = {2021},
month = {05},
pages = {},
title = {{The MRS UAV System: Pushing the Frontiers of Reproducible Research, Real-world Deployment, and Education with Autonomous Unmanned Aerial Vehicles}},
volume = {102},
journal = {Journal of Intelligent \& Robotic Systems},
doi = {10.1007/s10846-021-01383-5}
}

@INPROCEEDINGS{lam2010mpcc,
  author={Lam, Denise and Manzie, Chris and Good, Malcolm},
  booktitle={49th IEEE Conference on Decision and Control (CDC)}, 
  title={{Model predictive contouring control}}, 
  year={2010},
  volume={},
  number={},
  pages={6137-6142},
  doi={10.1109/CDC.2010.5717042}}

\end{document}